\documentclass{article}

\usepackage{arxiv}

\usepackage[utf8]{inputenc} % allow utf-8 input
\usepackage[T1]{fontenc}    % use 8-bit T1 fonts
\usepackage{hyperref}       % hyperlinks
\usepackage{url}            % simple URL typesetting
\usepackage{booktabs}       % professional-quality tables
\usepackage{amsfonts}       % blackboard math symbols
\usepackage{nicefrac}       % compact symbols for 1/2, etc.
\usepackage{microtype}      % microtypography
\usepackage{lipsum}		% Can be removed after putting your text content
\usepackage{amsmath}
\usepackage{graphicx}
\usepackage{natbib}
\usepackage{doi}

\usepackage{acronym}
\acrodef{RMS}[RMS]{Root mean square}
\acrodef{DBH}[DBH]{Diameter at Breast Height}
\acrodef{OKS}[OKS]{Object Keypoint Similarity}
\acrodef{FCNN}[FCNN]{fully-convolutional neural network}
\acrodef{RPN}[RPN]{Region Proposal Network}
\acrodef{RoI}[RoI]{regions of interests}
\acrodef{IoU}[IoU]{intersection over union}
\acrodef{TP}[TP]{true positive}
\acrodefplural{TP}{true positives}
\acrodef{FP}{false positive}
\acrodefplural{FP}{false positives}
\acrodef{FN}{false negative}
\acrodefplural{FN}{false negatives}
\acrodef{AI}[AI]{Artificial Intelligence}
\acrodef{AP}[AP]{Average Precision}
\acrodef{AR}[AR]{Average Recall}
\acrodef{S-LOAM}[S-LOAM]{Semantic LOAM}
\acrodef{IBC}[IBC]{Intelligent Boom Control}
\acrodef{ICAS}[ICAS]{intelligent computer assisted support}
\acrodef{fps}[fps]{frame per second}
\acrodef{auc}[AUC]{area-under-curve}
\acrodef{SGD}[SGD]{stochastic gradient descent}
\acrodef{CNN}[CNN]{Convolutional Neural Network}
\acrodef{IID}[IID]{Independent and Identically Distributed}
\acrodef{FLOP}[FLOP]{Floating Point Operation}

\usepackage{xspace}
\newcommand{\PortugalDB}{\textsc{Portugal}\xspace}
\newcommand{\CanaTreeDB}{\textsc{CanaTree100}\xspace}
\newcommand{\SynthTreeDB}{\textsc{SynthTree43k}\xspace}

\usepackage{array}
\usepackage{multirow}
\usepackage{booktabs}
\usepackage{makecell} % for bold lines
\usepackage{etoolbox}

\RequirePackage[online]{threeparttable} % For table notes
\let\TPTnoteSettingsOrig\TPTnoteSettings
\def\TPTnoteSettings{\TPTnoteSettingsOrig\scriptsize}
\RequirePackage{endnotes}
\newcommand*{\printendnotes}{%
	\def\enoteformat{\rightskip=0pt \leftskip=0pt \parindent=0pt \makeenmark}
	\def\makeenmark{\bgroup\bfseries\theenmark\enspace\egroup}
	\theendnotes
}

\title{Tree Detection and Diameter Estimation \\ Based on Deep Learning}

\author{ \href{https://orcid.org/0000-0002-8222-1266}{\includegraphics[scale=0.06]{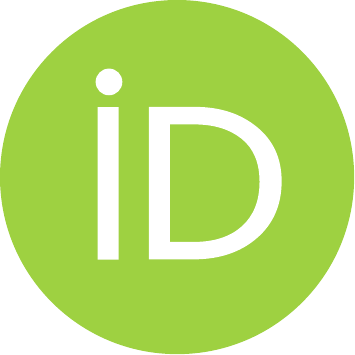}\hspace{1mm}Vincent Grondin}\thanks{vincent.grondin.2@ulaval.ca} \\
	Department of Computer Science\\
	Laval University\\
	\And
	{\hspace{1mm}Jean-Michel Fortin} \\
	Department of Computer Science\\
	Laval University\\
	\And
	\href{https://orcid.org/0000-0003-1288-2744}{\includegraphics[scale=0.06]{orcid.pdf}\hspace{1mm}François Pomerleau} \\
	Department of Computer Science\\
	Laval University\\
	\And
	\href{https://orcid.org/0000-0002-7520-8290}{\includegraphics[scale=0.06]{orcid.pdf}\hspace{1mm}Philippe Giguère} \\
	Department of Computer Science\\
	Laval University\\
}

\renewcommand{\shorttitle}{Tree Detection and Diameter Estimation Based on Deep Learning}

\hypersetup{
pdftitle={A template for the arxiv style},
pdfsubject={q-bio.NC, q-bio.QM},
pdfauthor={David S.~Hippocampus, Elias D.~Striatum},
pdfkeywords={First keyword, Second keyword, More},
}

\begin{document}
\maketitle

\begin{abstract}
	\looseness-1Tree perception is an essential building block toward autonomous forestry operations.
	Current developments generally consider input data from lidar sensors to solve forest navigation, tree detection and diameter estimation problems. Whereas cameras paired with deep learning algorithms usually address species classification or forest anomaly detection. 
	In either of these cases, data unavailability and forest diversity restrain deep learning developments for autonomous systems.
	So, we propose two densely annotated image datasets --- 43\,k synthetic, 100 real --- for bounding box, segmentation mask and keypoint detections to assess the potential of vision-based methods.
	Deep neural network models trained on our datasets achieve a precision of 90.4\,\% for tree detection, 87.2\,\% for tree segmentation, and centimeter accurate keypoint estimations.
	We measure our models' generalizability when testing it on other forest datasets, and their scalability with different dataset sizes and architectural improvements. Overall, the experimental results offer promising avenues toward autonomous tree felling operations and other applied forestry problems. The datasets and pre-trained models in this article are publicly available on \href{https://github.com/norlab-ulaval/PercepTreeV1}{GitHub} (https://github.com/norlab-ulaval/PercepTreeV1).\vskip1pt
\end{abstract}

% keywords can be removed

\section{Introduction}
\begin{figure}[h!]
		\centering{\includegraphics[width=0.8\linewidth]{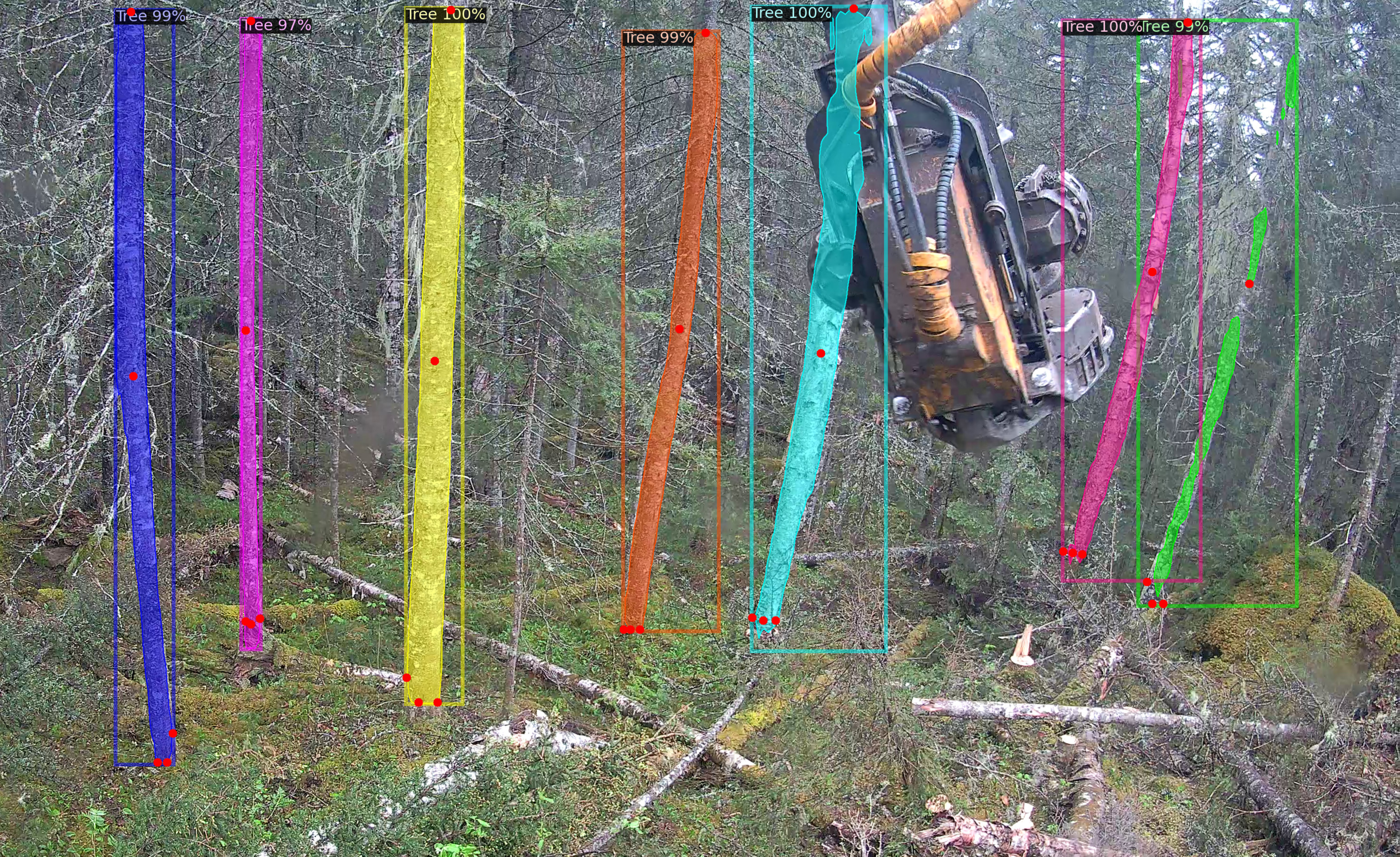}}
		%\vbox{\fbox{\hbox to 20pc{\bfseries\hfil FPO\hfil}}}
		\caption{
			Prediction from our vision-based model.
			The image is from an operator's point of view during felling operations in a Canadian forest. Each detected tree is shown in a different color and provides a bounding box, a segmentation mask and keypoint information.
		}
		\label{fig:dashcam_detections}
	\end{figure}

    % What is the problem?
	Heavy forest machinery can function at a capacity that is well above its operator's sustainable workload~\citep{ringdahl2011automation, parker2016robotics}.
	Hence, automation could increase operator efficiency while reducing mental fatigue~\citep{ ortiz2014increasing, visser2021automation}.
	% The enabling step to automation is tree perception~\citep{lindroos2015estimating}.
	% Ideally, the latter requires a method that is cost-effective and robust enough to perform in various forest conditions, yet few efforts take this into account.
	A technology driver behind advancements in automation is machine learning, which is an established approach to advance particular fields~\citep{roy2021machine}.
	Therefore, applying learning methods to a problem set specifically to the field of forestry, such as harvesting (Figure~\ref{fig:dashcam_detections}), forwarding, or navigation, will play an important role in the future of forestry automation. 
	However,  forests are harsh, unstructured outdoor environments~\citep{thorpe2001field}.
	As an environment becomes less structured, its complexity increases beyond the typical well-defined boundaries and narrow operating range established in structured environments, possibly leading to operational shortcomings of autonomous systems~\citep{roy2021machine}.
	
	% Why is it interesting and important?
	Autonomous systems, like humans, heavily rely on the perception of their surrounding environment to make decisions.
	While perception systems are adequate for clearly stated tasks and well-defined objects, general object recognition is still difficult.
	From a forestry standpoint, the capacity to quickly and reliably detect trees in various situations remains an important challenge for the development of multiple applications.
	For instance, tree selection in thinning operations is notably influenced by variable tree parameters such as stems and crowns density, height, diameter and species~\citep{nurminen2006time}, making the tree appearance exceedingly diverse. % vary wildly 
	As of now, the human operator's visual capacity is the only one capable of estimating these parameters and making informed decision during forestry operations.
	
	%\textbf{Why is it hard? (poor generalization performances)}
	However, replicating in machines the human perception system's effectiveness at distinguishing objects in different environments, and under a variety of conditions such as low illumination, color differences, and occlusions~\citep{padilla2021comparative}, is an active area of research.
	In computer vision, the most successful perception methods use images combined with deep neural networks and supervised learning~\citep{diez2021deep}. % might need a citation here
	As a data-centric method, supervised deep learning leverage large quantities of annotated images to discover intricate structures in high-dimensional pixel arrays~\citep{lecun2015deep}.
	An important limitation, although not unique to deep learning, is that any variation not learned within a training set can hinder performances and cause out-of-distribution generalization errors.
	This is particularly true for real-world applications, where the physical world is a large and rich source of information for training that it can't be seized by any dataset or even simulated worlds~\citep{roy2021machine}.
	In this context, a central challenge for perception systems is to generalize across a range of unforeseen and changing forest environments. % which involves reporting performances outside the training distribution.
	% Without capturing all the intricacies of the real-world, large datasets have proven useful ...
	Parallel to generalization concerns, the lack of resources to create a large dataset limits the adoption of deep learning for many problems~\citep{oliver2018realistic}, including tree detection.
	
	%\textbf{Why hasn't it been solved before (lots of data needed)}
	Creating large annotated datasets typically require great human effort and financial expenditure~\citep{oliver2018realistic}. 
	For instance, \citet{cheng2021pointly}'s analysis of human annotation time on COCO~\citep{lin2014microsoft} measures an average annotation time of 
	%(28.8\,s for category labeling, instance spotting 14.4\,s and 79.2\,s for segmentation mask annotation) 
	122.4\,s per instance.
	For a dataset with a size similar to COCO, this means thousands of annotation hours.
	One way to reduce these costs is to create a synthetic dataset using a simulator that can simultaneously generate and annotate images~\citep{james2019sim}.
	Provided that the perception system is pre-trained on a large synthetic dataset, one can achieve reasonably successful domain transfer (i.e. transfer learning) by fine-tuning the model on a small set of real data~\citep{oliver2018realistic}. 
	This would alleviate the under generalization issue deep learning models have when facing the domain shift between synthetic and real images.
	
	\begin{figure*}[h!]
		\centering{\includegraphics[width=1.0\linewidth]{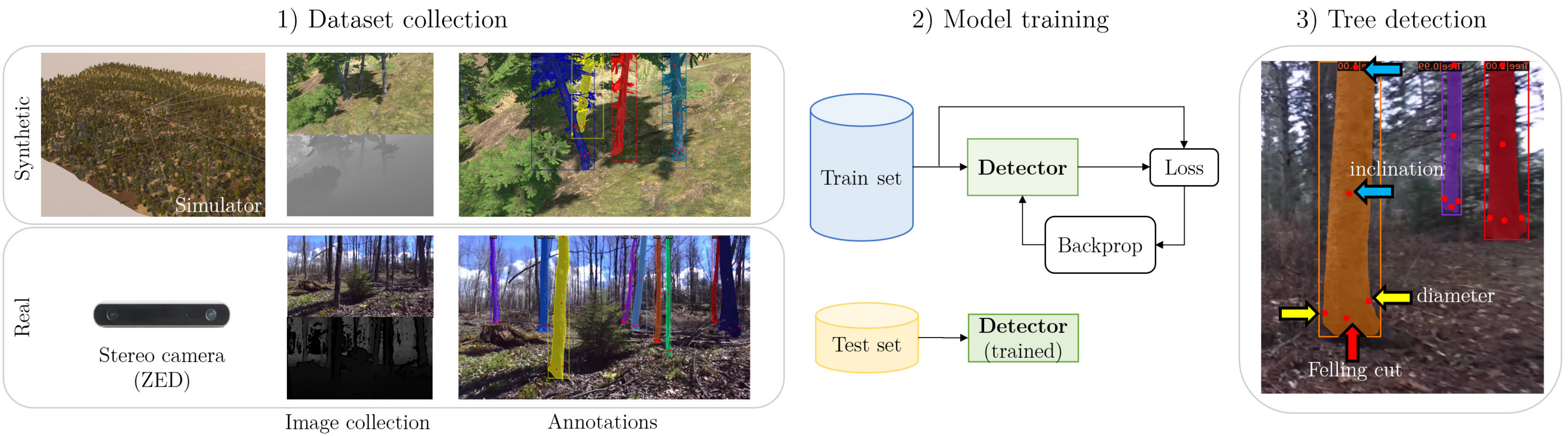}}
		\caption{
			Workflow of the proposed tree detector. 1) We create and annotate two forest datasets -- one synthetic and one real --.
				2) Using these datasets, object detector models are trained and tested.
				3) The model can detect, segment and estimate the felling cut, diameter and inclination.}
		\label{fig:worklfow}
	\end{figure*}
	
	%\textbf{What are the key components of my approach and results?}
	Based on the previous considerations, we propose two novel datasets to train common object detectors for tree perception, as well as felling cut, diameter and inclination estimation (see Figure~\ref{fig:worklfow}).
	To do so, we create a large synthetic dataset with software-generated annotations.
	We also generate a smaller dataset of real images that have been hand-annotated. 
	These datasets contain a vast variety of images and include class category, bounding box, segmentation mask and keypoint annotations. 
	Thereby, the perception problem is framed as one of instance segmentation to leverage the Mask R-CNN~\citep{he2017mask} framework and its derivative architecture, Cascade Mask R-CNN~\citep{cai2019cascade}. 
	Since these models have modular components that can be easily changed, this facilitates the integration of state-of-the-art architectures as they come out.
	In our experiments, we compare architectures and backbones to identify the most promising model. 
	Furthermore, limitations of our models are quantitatively evaluated by their capability to generalize on another dataset.
	We further analyze the benefits of synthetic images, detection reliability in presence of occlusion, the impact of train set size, and explore possible improvements for keypoint accuracy.
	Together, these experiments focus on discovering practical insights relevant to deep learning for autonomous tree detection.
	
	In short, our contributions are:
	%\textbf{Final paragraph or subsection: "Summary of Contributions"}
	
	\begin{itemize}
		\item Two annotated datasets of synthetic and real RGB-D images in natural forest environments are made publicly available;
		\item A performance comparison between different backbones and model architectures for tree detection; and
		\item Modified object detectors to estimate the felling cut, diameter and inclination for each tree.
	\end{itemize}

\section{Related Work}
\label{section:relatedWork}
	
	This literature survey is an overview of important advances in forestry automation.
	More specifically, we present works related to robotic applications in forests at a ground level, mainly towards autonomous forwarding and tree felling.
	Then, we cover works related to vision-based methods and expose how individual tree detection remains a crucial open problem.
	
	\subsection{Robotics in Forested Environments} 
	\label{sotaA}
    	According to \citet{ringdahl2011automation}, log transportation (i.e., forwarding) is the most likely operation to be automated, followed by tree extraction (i.e., tree felling).
    	An early proof-of-concept for forwarding automation is proposed by~\citet{hellstrom2009autonomous} in 2009.
    	They use a simulated environment where the vehicle moves along a path by using a path-search mechanism that generates feasible and obstacle-free 10~\,m long path segments.
    	In real-time, obstacle detection (i.e. trees) is attempted by the vehicle’s laser scanner sensors and an occupancy grid, but it is not clear if its been satisfactorily resolved.
    	More recently, \citet{jelavic2021towards} demonstrate a fully autonomous machine for tree felling operations.
    	Using a lidar sensor as its primary sensing modality, the automated system performs mapping, localization, planning, control and tool positioning. 
    	Tree detection is based on a geometry algorithm applied on 6\,m $\times$ 6\,m patches from a point cloud map.
    	During tests performed in a forest environment, the authors report successful detections in presence of vegetation, but the algorithm produces false negatives under heavy clutters. 
    	Also, the authors underscore that geometry-based detection harbor safety concerns, as the algorithm cannot distinguish between a standing person and a tree trunk.
    	
    	Apart from autonomous operations, tree detection from point clouds is used in other research experiments.
    	\citet{zhang2019rubber} extract the trunk's central point from clustered point clouds using a Gauss–Newton method. 
    	For their experiment they use a robot navigating in a structured, evenly spaced forest plantation.
    	In these conditions, they obtain a tree localization error slightly above 10\,cm (\ac{RMS}, as well as a sub-cm precision for the radius (\ac{RMS} = 0.66\,cm), rendering the method eligible to carry out autonomous tree harvesting.
    	In a similar way, ~\citet{pierzchala2018mapping} perform tree localization and \ac{DBH} estimation in a semi-structured, little undergrowth and flat terrain forest using graph-SLAM. 
    	They achieve a mean \ac{DBH} estimation error of 2\,cm and a tree position error of 4.76\,cm.
    	\citet{tremblay2020automatic} obtain comparable results in more challenging forests.
    	Based on a cylinder fitting method, they report a \ac{DBH} error of 2.04\,cm in mature forests with well-spaced trees and visible trunks, compared to 3.45\,cm error in unruly forest conditions.
    	\citet{liang2014use} proposed a photogrammetric approach for 3D imaging, known as Structure from Motion, and achieve a 2.39\,cm \ac{RMS} error on \ac{DBH} estimation of individual trees, which is acceptable for practical applications. 
    	The authors achieve similar results when using a terrestrial lidar~\citep{liang2014use}, which was used as a benchmark.
    	
    	% Link with my work
    	In our work, we aim for reliable tree detection and diameter estimation, but also go one step further by extracting relevant harvesting information such as tree felling cut location and inclination. 
    	This is an important contribution, since it would provide an end-to-end perception method ready for tree felling automation.

    \subsection{Vision-based deep learning} 
    \label{sotaB}
    	Limited work is conducted with deep learning to perform vision-based tree detection.
    	An image-based trunk detection system for forestry mobile robotics is presented in~\citet{da2021visible}. 
    	Using a mix of visible and thermal images, they create a dataset composed of 2895 images extracted from video sequences that contain bounding box annotations to detect the presence of tree trunks in images.
    	They train five different one-shot detectors on their dataset and achieve 89.84\,\% \ac{AP}, and 89.37\,\% F1 score.
    	Their dataset only include bounding box annotation, and exclude segmentation masks to precisely delimit pixels corresponding to trunks and keypoint annotations for each tree. 
    	Meanwhile, \citet{wang2019individual} use Faster R-CNN to detect tree trunks in depth images. 
    	These depth images are based on above-ground point cloud data, where the point clouds are extracted, voxelized and projected on a 2D plane to form an image.
    	When trained on 802 and tested on 359 of these images, they achieve more than 90\,\% of detection accuracy. 
    	\citet{itakura2020automatic} propose an occlusion-aware R-CNN for detecting trees in urban environments from reconstructed 3D images captured by a 360 spherical camera.
    	Using YOLOv2~\citep{redmon2016yolo9000}, they achieved 88\,\% precision, 100\,\% recall, and can estimate diameter with less than 3\,cm of error. 
    	Again, the stark difference between urban and natural forest environments render this dataset inoperative for forestry operations. 
    	
    	Deep learning is also  used to classify trees and estimate stock volume.
    	\citet{liu2019classification} train a U-Net~\citep{UNet2015} for instance segmentation on more than 3\,k images, and then process the segmentation masks with a non-linear mixed effect model to estimate stock volumes, achieving 97.25\,\% precision, 95.68\,\% recall, 96.03\,\% classification accuracy and $\mathrm{30.54\,m^3/ha}$ stock volume \ac{RMS} error. 
    	Further, \citet{ostovar2019detection} tackle stump detection and classification in forests by training Faster R-CNN~\citep{FasterRCNN} on 800 images and then testing it on 200 images. 
    	They report precision and recall rates of 95\,\% and 80\,\%, respectively.
    	
    	A common theme among the aforementioned tasks is the obligation to detect individual trees first, which allow subsequent tasks, like stock volume estimation or species classification. 
    	Insofar, precision and recall rates reported for tree detection among the precedent works are relatively high, which does not seem to truly reflect its difficulty.
    	In this regard, a main contribution of our work is to measure the performance of deep learning on this specific task, and quantitatively demonstrate its limitations when deployed on out-of-distribution forest environments. 
    	% By releasing our dataset, others can also do the same. 

\section{Method}
\label{section:method}
	The methodology is divided into four sections.
	First, we detail our two datasets, \SynthTreeDB and \CanaTreeDB.
	Then, object detection models are presented, followed by training parameters.
	Finally, we describe the evaluation metrics used in the experiments.
	
	\subsection{Datasets}
	\label{subsection:Datasets}
    	One of the key issues restraining the adoption of deep learning approaches in the forestry domain is the limited availability of annotated datasets dedicated to forest applications.
    	Although this research area has disruptive potential toward automation, no existing dataset contains enough annotation to train a tree detector with exact pixel location of the trunk, felling cut, diameter and inclination. 
    	A major advantage of simulation is the ability to fully control the data generation pipeline, which ensures reduced costs, and infinite variety and quantity~\citep{gaidon2016virtual}. 
    	Therefore, we propose to employ simulated virtual forests as a proxy for our tree detection tasks.
    	As it has already been shown~\citep{gaidon2016virtual, cabon2020virtual}, pre-training deep neural networks on virtual data improve object detection performances. So, we created a large novel synthetic dataset to partially fill the data gap in forestry.

\begin{figure}[h!]
     		\centering{\includegraphics[width=.8\linewidth]{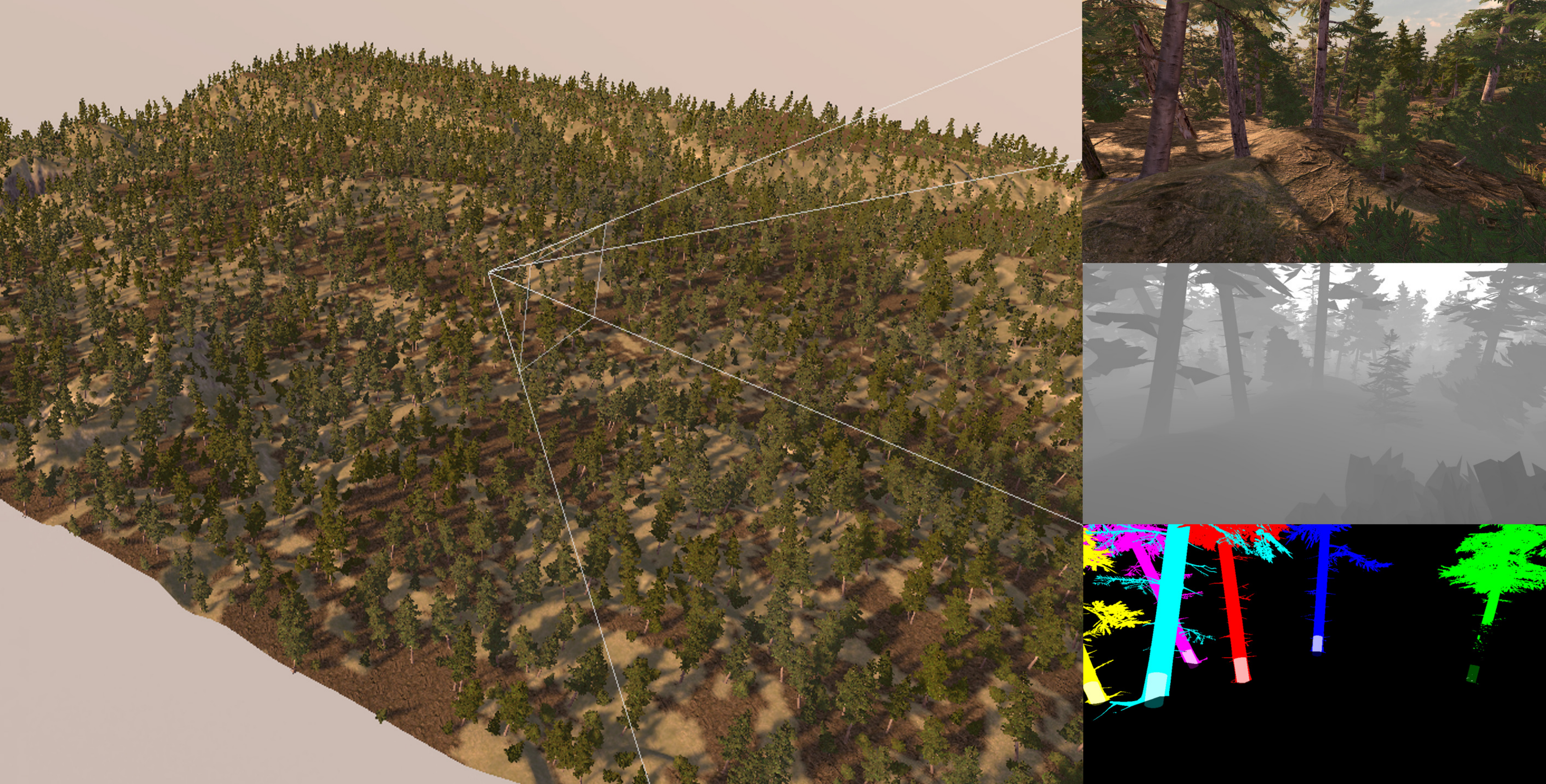}}
     		%\vbox{\fbox{\hbox to 20pc{\bfseries\hfil FPO\hfil}}}
     		\caption{The \SynthTreeDB dataset. \textbf{Left} a general view of a simulated scene where the camera point of view is represented by white lines. This scene has around 1500 tree instances, of which we sample 750 images. Along with a morning light effect, and the terrain texture follows an altitude rule where the soil low and the grass is high. The scene also contains scrubs, grass and branches. \textbf{Right} a sample RGB frame, its depth frame, and its segmentation mask annotations. Individual trees are in different colors, and the tree base is slightly paler.\label{fig:sim_world}}
     	\end{figure}

    	Named \SynthTreeDB, the dataset is a collection of 43\,k synthetic images rendered in a simulated environment using the Unity\footnote{https://unity.com} game engine.
    	The image resolution is set to 1280\,$\times$\,720 pixels.
    	To build realistic forest scenes (see Figure~\ref{fig:sim_world}), we employ Gaia\footnote{https://assetstore.unity.com/packages/tools/terrain/gaia-2-terrain-scene-generator-42618}, a Unity asset, to efficiently terraform landscapes, texture terrain, and place objects models. 
    	Forest types described by \citet{diez2021deep} --- plantation, well-managed and natural forests --- are taken into account by controlling trees spawn density.
    	Then, six basic 3D tree models from Nature Manufacture\footnote{https://naturemanufacture.com/} (from fir and beech species) are modified by changing bark and foliage textures to create 17 different trees. 
    	During scene generations, each tree is placed according to spawn rules about the terrain altitude, slope and object density in a given area. 
    		When spawned, a tree is assigned a random width and height for increased appearance diversity.
    	Other objects and conditions commonly found in forests are also included in this simulated environment :
    	\begin{itemize}
    		\item grass, stumps, scrubs and branches;
    		\item lighting conditions (i.e., morning, daytime, evening, and night light); and
    		\item weather effects related to fog, snow, and rain.
    	\end{itemize}
    	
    	We sample images from the scene by arbitrarily moving the camera from tree to tree with random roll, pitch and yaw angle within a certain range to mimic small changes in camera pose caused by uneven terrain navigation.
    	For each generated scene, 200 to 1000 images are sampled depending on the number of trees in the scene.
    	This is done in an effort to reduce the odds of the same tree appearing in multiple images.  
    	% We also abstain from annotating trees that are too far away, for the same reason. 
    	For the annotation method, a tree is only annotated if within the typical feller crane's reach, 10\,m, which in our case is its distance from the camera.
    	
\begin{figure}[h!]
     		\centering
     		\includegraphics[ width=1.0\linewidth]{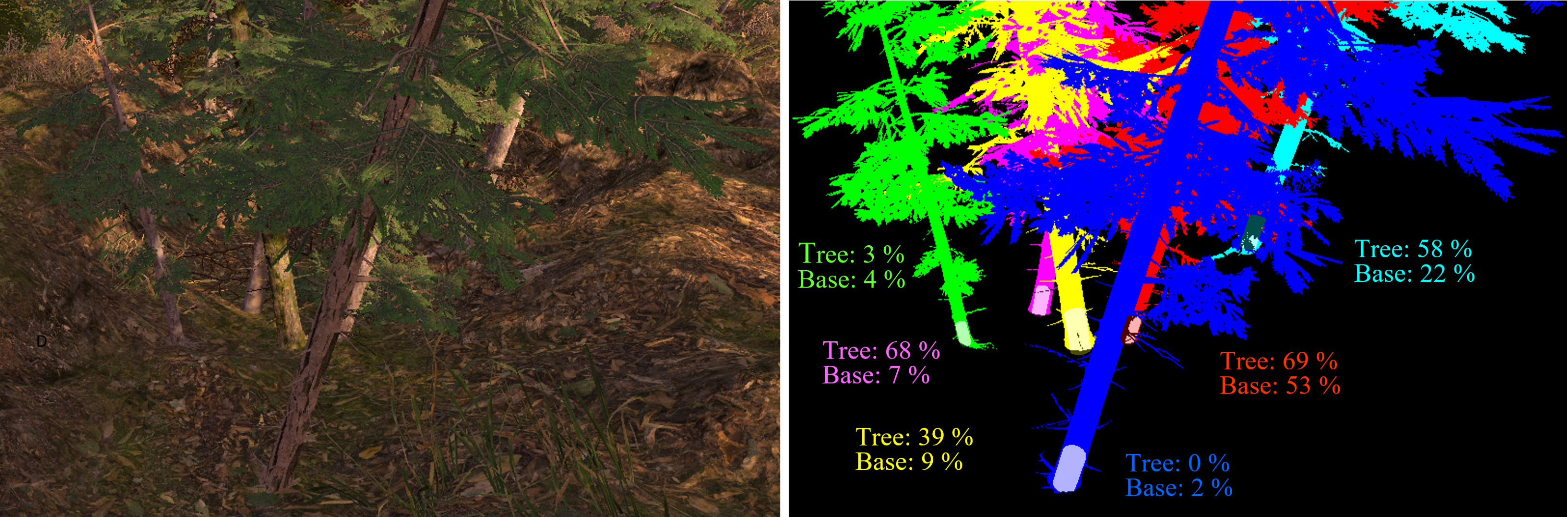}
     		\caption{\textbf{Left} RGB synthetic image and \textbf{right} its segmentation mask (later on fitted to the trunk). Each tree has a segmentation mask denoted by a different color. Lighter colored areas correspond to the tree base. Occlusion is measured based on the percentage of visible tree pixels visible the image frame.}
     		\label{fig:occlusion_illustrated}
     	\end{figure}

    	Bounding box annotation is confined to the trunk to omit the tree crown foliage, and this differs from the urban tree dataset in~\citet{itakura2020automatic}, where bounding box annotations contain entire trees.
    	This decision is motivated by the fact that 1) tree density in natural forests makes them more prone to overlapping error than in urban environments, and 2) tree felling primarily requires information about the trunk.
    	Contrary to the works of~\citet{itakura2020automatic} and~\citet{da2021visible}, we add segmentation mask annotations, and provide pixel information.
    	We equally leverage the simulator to compute the percentage of tree occlusion from the camera's point of view (see Figure~\ref{fig:occlusion_illustrated}), which is used later on to quantify the model's robustness with regard to occlusion.
    	The percentage of occlusion takes into account occlusion from grass, other trees and the landscape.
    	Moreover, we compute the entire tree visibility and ignore tree instances if less than 30\,\% of their pixels are visible. 
    	This threshold was set to restrain training on instances with too much occlusion.
    	Such instances can be misleading, causing the detection thresholds to be lowered, which in turn increases the objects' context influence, and leads to \ac{FP} in regions with no object~\citep{wang2020robust}.
    	Lastly, we locate five keypoints on each tree to capture cues relevant to harvesting.
    	These keypoints correspond to the tree's felling cut location, diameter and inclination.
    	Our simulator setup can generate these annotated synthetic images at a rate of approximately 20 frames/minute, for we consider RGB and depth images as one frame. 
    	Our final synthetic dataset contains over 43\,k RGB-D images, and more than 162\,k annotated trees.
    	
    	Our real dataset, referred to as the \CanaTreeDB dataset, was collected from public, private, and commercial forests in Quebec, Canada.
    	The images were taken from June 2020 to April 2021, between the times of 9\,am and 6\,pm. 
    	Tree species commonly found in these forests include fir, spruce, pine, birch and maple.
    	Our images are hand-annotated with the COCO-Annotator~\citep{cocoannotator}.
    	During annotation, we manually draw the segmentation mask on the trunk only, fit bounding box to it and place keypoints.
    	As opposed to the annotations in the synthetic dataset, we omit tree branches during segmentation annotation (that fit inside the bounding box), since they are complex to annotate and we have no use for it in forestry operations.
    	The felling cut and diameter keypoints are approximately placed 10\,cm above ground.
    	For the two inclination keypoints along the tree trunk, contrary to the simulator that placed them at 90\,cm intervals starting from the felling cut, we decided to place one in the middle and one on top.
    	
    	Overall, \CanaTreeDB provides 100 RGB and 100 depth images, as well as more than 920 annotated trees. 
    	The images are garnered from 33 videos captured by a hand-held ZED stereo camera, and have HD of 1280\,$\times$\,720{@}60\,fps resolution. Like the synthetic dataset, the RGB and depth images are considered as one frame. 
    	We manually select approximately three images per video ($<1$\% of the video).
    	We subjectively select images that are challenging and rich in information.
    	Because of this very low sampling rate, the correlation between samples is minimized, but \CanaTreeDB is significantly smaller than other analogous datasets~\citep{liu2019classification, da2021visible} that employ much higher sampling rates from videos.
    	Although using a high frame rate allows for efficient image annotation due to temporal correlation, it might result in impoverished visual diversity, which is inefficient when training deep networks.
    	
    	So far, our datasets are limited to the situation where train and test sets come from forests of similar ecosystems (deciduous, mixed or boreal forests from Quebec).
    	Since forest machinery operates in a variety of areas, there is a fundamental concern about our tree detector's ability to generalize to other regions, without fine-tuning.
    	In an effort to address this, we use the trunk detection dataset from \citet{da2021visible}, which we refer to as the \PortugalDB dataset.
    	This should help us measure the extent at which deep learning models trained on our datasets generalize to a different domain
    	The domain transfer experiments in Section \textit{Domain transfer} only measure performances for bounding box detections, as the dataset only contains bounding box annotations.
    	This dataset is composed of 2690 RGB images collected from three different forest sites in Portugal, to which the authors applied eight unique data augmentations to every image, increasing their dataset size to a total of 24\,k images. % (8+1) x 2690 (includes the original one)
    	% This lead to highly correlated samples.
    	For a fair comparison, we use the same train, validation and test split as \citet{da2021visible}.

	\subsection{Model architectures}
	\label{subsection:ModelArchitectures}
\begin{figure}[h!]
    		\centering{\includegraphics[width=.6\linewidth]{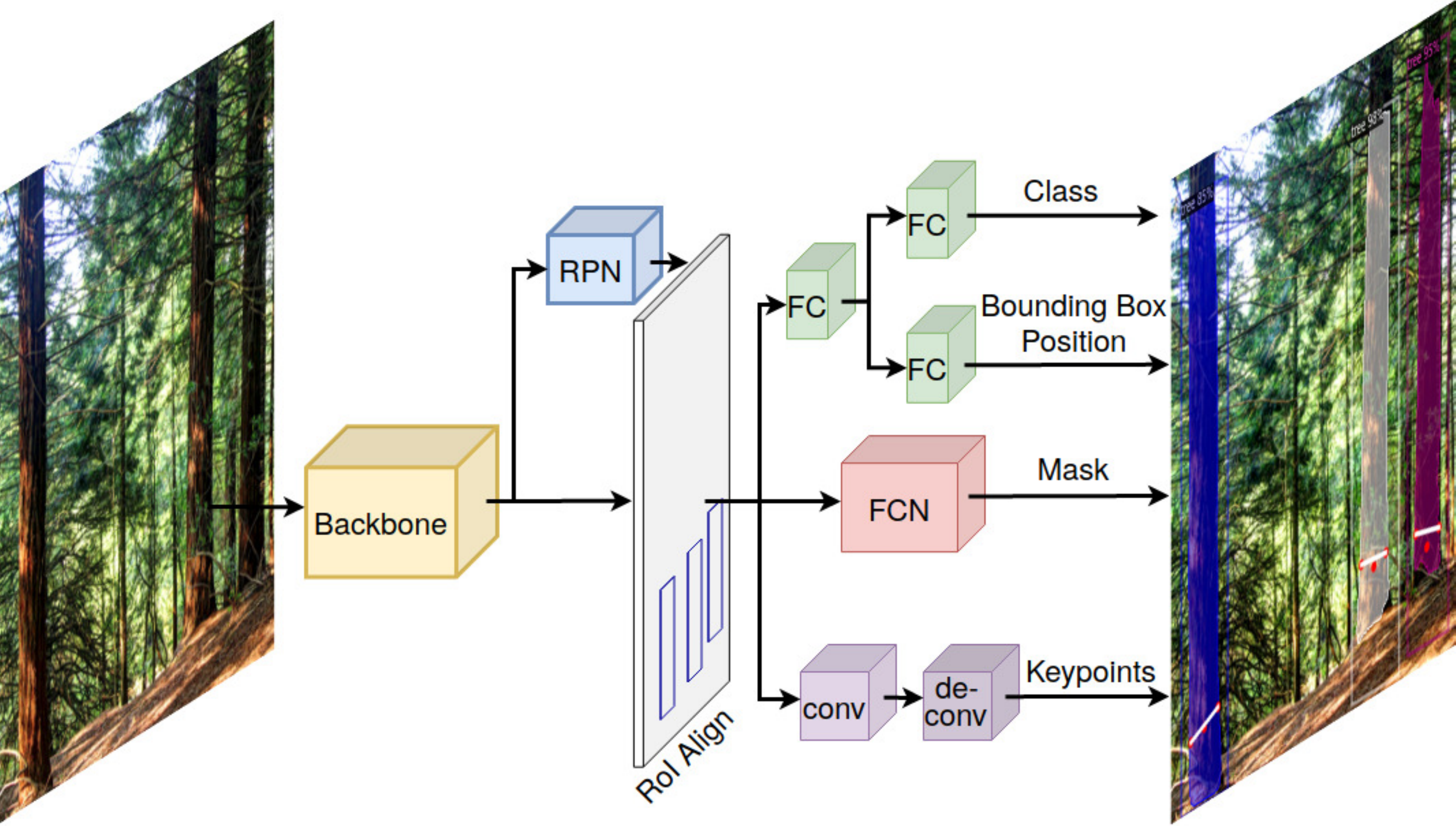}}
    		\caption{Model architecture of Mask R-CNN with a keypoint branch. 
    			For each input image, the backbone extracts distinctive features to predict \ac{RoI}s and the three network heads.
    			The first head (green) predicts the class and bounding box, the second head (red) predicts the segmentation mask, and the third head (purple) predicts the keypoints.
    			This unified architecture enables end-to-end training without any post-processing.\label{fig:model_architecture}}
    	\end{figure}

    	Two different model architectures are used in the experiments --- Mask R-CNN and Cascade Mask R-CNN.
    	These models have two-stages, in which candidate bounding boxes are proposed by a \ac{RPN}, and then three network heads individually predict a bounding box, segmentation mask and keypoints (see Figure~\ref{fig:model_architecture}). 
    	
    	In the first stage, images are first processed by the backbone to extract distinctive object features.
    	To do so, we experiment with three backbone architectures. 
    	For instance, we use ResNeXt~\citep{xie2017aggregated}, a \ac{CNN}-based backbone well suited for the grid-structure of images.
    	For a long time, \ac{CNN}s have dominated computer vision with their intrinsic capability to impose regularizing priors for object detection like stationarity, locality and compositionality~\citep{bronstein2017geometric}.
    	Image features can also be extracted using transformer-based backbones such as Swin~\citep{liu2021swin}.
    	Beyond being state-of-the-art in many domains, transformers are able to model long-range dependencies in data through a self-attention mechanism.
    	We conjecture that this mechanism might be beneficial for forest targeted application, since short-range information is often obfuscated by overlapping trees or occlusion.
    	The Swin transformer also employs a shifted-window strategy for self-attention, which improves efficiency at distinct object feature extraction, important for real-time operations.
    	The third and last feature extraction method we experiment with is CBNetV2~\citep{liang2021cbnetv2}.
    	Without introducing a new feature extraction paradigm, this method uses dual backbones (DB).
    	Contrary to simple network deepening or widening, the CBNetV2 backbone architecture integrates high- and low-level features of multiple identical backbones that are connected through composite connections. 
    	It can be adapted to various backbones (i.e. CNN-based, Transformer-based) as well as head designs of most mainstream detectors.
    	Table~\ref{tab:hyperparameter_train} presents the complete list of backbones and number of trainable parameters used in our experiments. 
    	
    	The last step of the first stage is to pass the extracted feature to the \ac{RPN} (the blue module in Figure~\ref{fig:model_architecture}), which identifies a set of regions that could contain objects. %which are then used by a downstream detector module to localize and classify objects.
    	In doing so, the \ac{RPN} evaluates the objectness of each pre-defined axis-aligned anchor box corresponding to three area-scales (8, 16 and 32) and three aspect ratios (0.5, 1.0, and 2.0) to generate region proposals.
    	
    	Then, during the second stage, RoiAlign~\citep{he2017mask} extracts small feature maps for each region proposal using a bilinear interpolation to map the feature vectors extracted by the backbone, into a $7\times7$ input feature vector.
    	Subsequently, feature vectors from each \ac{RoI} go through the three network heads.
    	The first head predicts the class and bounding box, the second head predicts the segmentation mask, and the third head predicts the keypoints.
    	The segmentation head performs semantic segmentation with an output resolution of 28 $\times$ 28.
    	The keypoint head accuracy is set at 56 $\times$ 56, because keypoint localization requires a higher resolution~\citep{he2017mask}.

	\subsection{Training}
	\label{subsection:TrainingParameters}
    	We use the PyTorch based library MMdetection~\citep{mmdetection} for model implementations. 
    	All backbone weights are initialized from their pre-trained version on COCO~\citep{lin2014microsoft} 2017, a large-scale dataset containing 80 object classes and 164\,k images. % https://www.tensorflow.org/datasets/catalog/coco
    	This pre-training phase is important to help regularize the networks~\citep{erhan2010does} and facilitate transfer learning from a source to a target domain~\citep{ mahajan2018exploring}.
    	\citet{hinterstoisser2018pre} found no significant gain to re-train the entire backbone, hence some backbone stages are frozen before training or fine-tuning the models. 
    	So, in our experiments, the first three (out of 4) stages are frozen in the Swin backbone, whereas in ResNeXt-101 is only frozen in the first (out of 5) stage. 
    	We use an initial learning rate of 1e-4 for pre-training and 5e-5 for fine-tuning, decreasing the rate by a factor of 10 at predetermined steps. 
    	The steps were obtained through a grid-search process by increments of 10.
    	Pre-training and fine-tuning is done on two \textit{NVIDIA A100} GPU (40 GB memory).
    	
    	Pre-training on \SynthTreeDB is conducted on three subsets --- a train set, a validation set, and a test set --- of 40\,k, 1\,k and 2\,k images, respectively.
    	At train time, we employ common data augmentation techniques (i.e. image resize, horizontal flip, sheer, and camera distortions), and these significantly improve model generalization.
    	The model processes images from the train set in batches of four, and it is optimized with AdamW and a weight decay of 0.05. 
    	This optimizer regularizes learning by using the weight decay.
    	As training progresses, the validation set monitors overfitting and performs early stopping.
    	
    	\begin{table}[hbt!]
    		\centering
    		\caption{Model parameters, \ac{FLOP}s and \ac{fps} on an NVIDIA RTX 3090.} \label{tab:hyperparameter_train}
    		\begin{threeparttable}
    			\begin{tabular}{c|c|ccc}
    				\toprule%
    				Model & Backbone & \#Params & FLOPs & fps \\
    				\midrule
    				\multirow{3}{*}{Mask} & X-101 & 63M & 238G & 12.5 \\ 
    				& Swin-T & 48M & 267G & 12.8  \\
    				& DB-Swin-T & 76M & 357G & 11.7 \\
    				\midrule
    				\multirow{3}{*}{Cascade} & X-101 & 101M & 819G &  7.2 \\ 
    				& Swin-S & 107M & 832G & 7.0\\
    				& DB-Swin-S & 156M & 1016G & 6.4\\
    				\bottomrule
    			\end{tabular}
    		\end{threeparttable}
    	\end{table}

    	% Once pre-trained on synthetic images, we proceed to fine-tune the network on real images.
    	Although models pre-trained on synthetic images can learn the tree appearance in simulation and transfer this knowledge for detection in the real world, their performance can be improved, sometimes significantly, when fine-tuned on real images\citep{gaidon2016virtual}.
    	This fine-tuning step accounts for the reality gap between the two domains (synthetic vs. real), and the minor differences that might exist between synthetic annotations and manual annotations, for instance the bounding box height and keypoint positions.
    	Therefore, transfer learning to real images is done by fine-tuning the model on \CanaTreeDB.
    	Due to its smaller size, we use a five-fold cross-validation scheme: 60 in the train set, 20 in the validation set, and 20 in the test set. 
    	Importantly, we did not segregate physical sites when doing the train, validation and test split.
    	This means that images share similar characteristics and lighting conditions in different splits.
    	However, the same trees are not present in more than one set.
    	Data augmentation is used for fine-tuning, but considering the small train set size, an image batch size of two is preferred.
    	The final results are reported on the test set without any data augmentation, as is commonly done.

	\subsection{Detection metrics}
	\label{subsection:DetectionMetrics}
    	The quality of our tree detector can be measured using two criteria: 1) its ability to find all ground-truth trees, while 2) identifying only relevant trees~\citep{padilla2021comparative}. 
    	For the first criterion, the detector is considered to have a high recall rate when there is very few \acp{FN}. 
    	Then, when evaluated by the second criterion, the detector is considered to have a high precision rate when there are very few false positives (FPs). 
    	In case an \ac{FP} or \ac{FN} happens during forestry operations, there is a risk of decreasing productivity or damaging equipment.
    	Detection results are reported using the standard COCO metrics: \acf{AP} and \ac{AR}.
    	In our experiments, we measure the bounding box predictions (\ac{AP}$^{bb}$, \ac{AR}$^{bb}$) and segmentation mask predictions (\ac{AP}$^{seg}$, \ac{AR}$^{seg}$)
    	
    	The \ac{AP} takes the weighted sum of precision at each \ac{IoU} threshold:
    	
    	\begin{equation}
    	   AP = \frac{1}{N}\sum_{r} p_\text{interp}(r), \\
    	\label{eq:AveragePrecision}
    	\end{equation}
    	where \ac{IoU} is defined as the ratio between the intersection and the union of the predictions and the ground truths; $p_\text{interp}(r)$ is the interpolated precision at each equally spaced recall level $r=\{0, 0.01, \dots, 1\}$, with $N=101$.
    	The \ac{AP} is computed following the standard intervals of \ac{IoU}\,=\,[0.5:0.05:0.95]. 
    	For clarity, we denote it \ac{AP}$_{50:95}$.
    	Similarly, we also report \ac{AP}$_{50}$ results computed for \ac{IoU}\,=\,0.5, since many previous works~\citep{da2021visible, ostovar2019detection, itakura2020automatic, zhang2019rubber} only report their results using 50\,\% minimum \ac{IoU}.
    	
    	The \ac{AR} measures the assertiveness of object detectors.
    	It is computed using the average of maximum recall across several \ac{IoU} thresholds $O$:
    	
    	\begin{equation}
    	AR = \frac{1}{O}\sum_{o=1}^{O} \max\limits_{k|Pr_{t(o)}(\tau(k)>0} \{Rc_{t(o)}(\tau(k)\}, \\
    	\label{eq:AveragerRecall}
    	\end{equation}
    	where $Pr_{t(o)}(\tau(k)$, $Rc_{t(o)}(\tau(k)$ are the precision $\times$ recall points for $\tau(k)$ confidence. 
    	In our experiments, we use the \ac{AR}$_{50:95}$, \ac{AR}$_{50}$.

\section{Results and Discussions}
\label{section:results}
	In order to evaluate our approach, we first test it on \CanaTreeDB. 
	After measuring the capability of our tree perception method on its source domain, we focus our attention on its generalization performances on another forest dataset.
	Finally, we conduct further studies quantifying \emph{i)} the impact of synthetic images during pre-training, \emph{ii)} model robustness to occlusion \emph{iii)} performance scalability as a function of dataset size, and \emph{iv)} keypoint architectural improvements.  

	\subsection{Main results}
	\label{subsection:MainResults}
\begin{figure*}[h!]
			\centering{\includegraphics[width=1.0\linewidth]{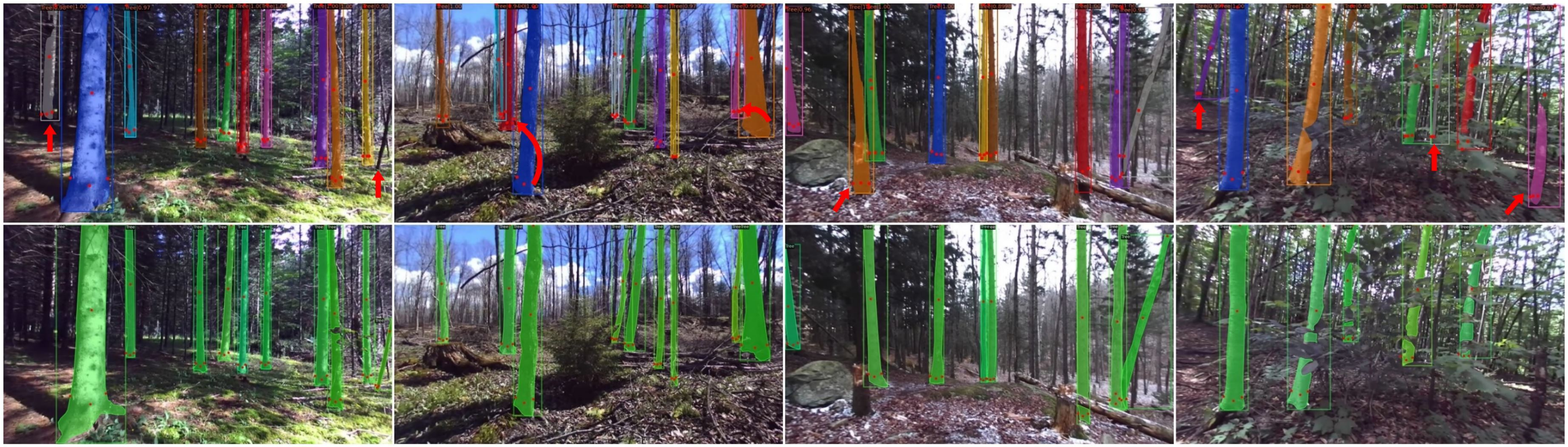}}
			%\vbox{\fbox{\hbox to 20pc{\bfseries\hfil FPO\hfil}}}
			\caption{\textbf{(Top row)} Detections are in comparison to their \textbf{(bottom row)} ground truths. 
				The straight arrows point out the missed and false detections. Missed detections are mostly caused by tree sizes, while false detections are often due to tree distance. 
				The curved arrows point out failed keypoint localizations commonly caused by overlapping trees; as a result, the keypoint is predicted on the wrong tree. 
				The prediction model (Cascade Mask R-CNN DB-Swin-S) is pre-trained on \SynthTreeDB, and fine-tuned on \CanaTreeDB.}
			\label{fig:detection_examples}
		\end{figure*}

	\textbf{Tree detection:}
		% The models' weights in this section are initialized on COCO, then pre-trained on \SynthTreeDB.
		% Having learned useful features from this pre-training phase, fine-tuning and evaluation is then performed on \CanaTreeDB.
		Tree detector performances based on different backbones and model architectures are presented in Table~\ref{tab:mainResults}.
		There is a consistent increase in detection results for Cascade Mask R-CNN models relative to Mask R-CNN models.
		Having several cascaded heads that progressively refine predictions improve results, and this is coherent with the results obtained by~\citet{cai2019cascade}.
		For backbones, transformer-based backbone (Swin) systematically outscore their \ac{CNN} counterpart.
		Using dual backbones (DB) further improves detection and segmentation results.
		Overall, the model with the most (156\,M) learnable parameters, Cascade Mask-RCNN DB-Swin-S, obtained a detection performance of 90.4\,\% \ac{AP}$^{bb}_{50}$ and 94.5\,\% \ac{AR}$^{bb}_{50}$.
		While these results are almost identical to the ones of Mask-RCNN DB-Swin-S --- 90.3\,\% \ac{AP}$^{bb}_{50}$ and 94.9\,\% \ac{AR}$^{bb}_{50}$ --- the Cascade model has a significant edge for all other detection metrics, notably gaining 3.8\,\% \ac{AP}$^{bb}_{50:95}$ and 1.5\,\% \ac{AR}$^{bb}_{50:90}$.
		Qualitative results can be observed in Figure~\ref{fig:detection_examples}.
		
		\begin{table*}[h!]
			\centering
			\caption{Detection performance on \CanaTreeDB. Keypoint errors relating to diameter (dia), felling cut (fc) and inclination (inc) are presented in metric units. Transformer backbones improve results, and dual-backbones consistently increase results for all tasks. Models are pre-trained on COCO and \SynthTreeDB.}
			\label{tab:mainResults}
			 \makebox[\linewidth]{
			 % Scale down table to \textwidth
			 \resizebox{16.4cm}{!}{%
				\begin{threeparttable}
				\begin{tabular}{cc|cccc|cccc|ccc}
					\toprule%
					\multirow{2}{*}{Model} & \multirow{2}{*}{Backbone} &  \multirow{2}{*}{AP$^{bb}_{50}$} & \multirow{2}{*}{AP$^{bb}_{50:95}$} & \multirow{2}{*}{AP$^{seg}_{50}$} & \multirow{2}{*}{AP$^{seg}_{50:95}$} & \multirow{2}{*}{AR$^{bb}_{50}$} & \multirow{2}{*}{AR$^{bb}_{50:95}$} & \multirow{2}{*}{AR$^{seg}_{50}$} & \multirow{2}{*}{AR$^{seg}_{50:95}$} & \multicolumn{3}{@{}c@{}}{Keypoint error} \\\cmidrule{11-13}
					&  &  &  &  &  &  &  &  &  & dia$_{(cm)}$ & fc$_{(cm)}$ & inc$_{(^{\circ})}$\\ 
					\midrule
					\multirow{3}{*}{Mask} & X-101 & 86.3$_{\pm 2.2}$ & 60.4$_{\pm 0.9}$ & 82.8$_{\pm 1.3}$ & 54.2$_{\pm 1.5}$ & 90.7$_{\pm 1.3}$ & 66.9$_{\pm 0.3}$ & 87.5$_{\pm 1.3}$ & 60.2$_{\pm 1.0}$ & 3.1 & 6.4 & 2.5 \\ 
					& Swin-T & 88.2$^{\uparrow1.9}_{\pm 1.2}$ & 59.6$^{\downarrow0.8}_{\pm 2.6}$ & 83.5$^{\uparrow0.7}_{\pm 1.6}$ & 55.8$^{\uparrow1.6}_{\pm 2.2}$ & 93.2$^{\uparrow2.5}_{\pm 1.3}$ & 67.1$^{\uparrow0.2}_{\pm 1.7}$ & 88.4$^{\uparrow0.9}_{\pm 2.3}$ & 61.8$^{\uparrow1.6}_{\pm 1.9}$ & 3.1$^{0.0}$ & 6.4$^{0.0}$ & 1.5$^{\downarrow1.0}$\\ 
					& DB-Swin-T & 90.3$^{\uparrow4.0}_{\pm 2.8}$ & 61.3$^{\uparrow0.9}_{\pm 1.4}$ & 86.3$^{\uparrow3.5}_{\pm 1.1}$ & 58.8$^{\uparrow4.6}_{\pm 2.0}$ & \textbf{94.9$^{\uparrow4.2}_{\pm 1.9}$} & 68.9$^{\uparrow2.0}_{\pm 1.2}$ & 91.1$^{\uparrow3.6}_{\pm 1.3}$ & 64.4$^{\uparrow4.2}_{\pm 1.3}$ & \textbf{2.8$^{\downarrow0.3}$} & 6.3$^{\downarrow0.1}$ & 1.5$^{\downarrow1.0}$ \\
					\midrule
					\multirow{3}{*}{Cascade} & X-101 & 88.1$_{\pm 3.4}$ & 61.7$_{\pm 1.7}$ & 83.1$_{\pm 1.4}$ &  55.9$_{\pm 1.2}$ & 93.7$_{\pm 3.6}$ & 70.1$_{\pm 1.5}$ & 87.2$_{\pm 1.4}$ & 61.3$_{\pm 1.0}$ & 3.6 & 7.0 & 1.8\\
					& Swin-S & 88.9$^{\uparrow0.8}_{\pm 2.4}$ & 61.6$^{\downarrow0.1}_{\pm 1.9}$ & 85.9$^{\uparrow2.8}_{\pm 1.5}$ & 57.5$^{\uparrow0.3}_{\pm 1.4}$ & 94.0$^{\uparrow3.2}_{\pm 2.6}$ & 68.7$^{\downarrow1.4}_{\pm 1.6}$ & 90.4$^{\uparrow3.2}_{\pm 1.5}$ & 62.9$^{\uparrow1.6}_{\pm 1.1}$ & 3.8$^{\uparrow0.2}$ & 6.4$^{\downarrow0.6}$ & 1.3$^{\downarrow0.5}$ \\ 
					& DB-Swin-S & \textbf{90.4$^{\uparrow2.3}_{\pm 2.4}$} & \textbf{64.1$^{\uparrow2.4}_{\pm 1.4}$} & \textbf{87.2$^{\uparrow4.1}_{\pm 1.7}$} & \textbf{60.0$^{\uparrow4.1}_{\pm 1.4}$} & 94.5$^{\uparrow0.8}_{\pm 1.9}$ & \textbf{70.4$^{\uparrow0.3}_{\pm 1.1}$} & \textbf{91.5$^{\uparrow4.3}_{\pm 1.4}$} & \textbf{65.2$^{\uparrow4.3}_{\pm 1.1}$} &3.6$^{0.0}$ & \textbf{6.1$^{\downarrow0.9}$} & \textbf{1.1$^{\downarrow0.7}$} \\
					\bottomrule
				\end{tabular}
				\begin{tablenotes}[para]
					\item[]Note: ${\uparrow, \downarrow}$: gain/decrease from the baseline model; $\pm$: standard deviation over the five folds.
				\end{tablenotes}
				\end{threeparttable}
			}
			}
		\end{table*}

	\textbf{Tree segmentation:}
		Pixel-level segmentation allows for a rich and detailed understanding of image content, which can play an important role in precisely delimiting the boundaries of individual trees \citep{liu2020deep}.
		Our best model, Cascade Mask R-CNN DB-Swin-S, achieves 87.2\,\% \ac{AP}$^{seg}_{50}$.
		One can also observe that models with high \ac{AP}$^{seg}$ performances achieve significantly better \ac{AP}$^{bb}$ and \ac{AR}$^{bb}$.
		While these findings could simply be attributed to the model's architecture, a previous study~\citep{he2017mask} demonstrated that increased bounding box and keypoint detections occur by learning features specific to segmentation.
		From an application standpoint, this is promising for stock volume estimation, where \citet{liu2019classification} successfully did it with a lower (86.0\,\%) \ac{AP}$^{seg}_{50}$ than us.
		Other applications, such as visual assistance for machine operators, can also benefit from pixel segmentation.
		
\begin{figure}[h!]
			\centering
			\centering{\includegraphics[width=0.6\linewidth]{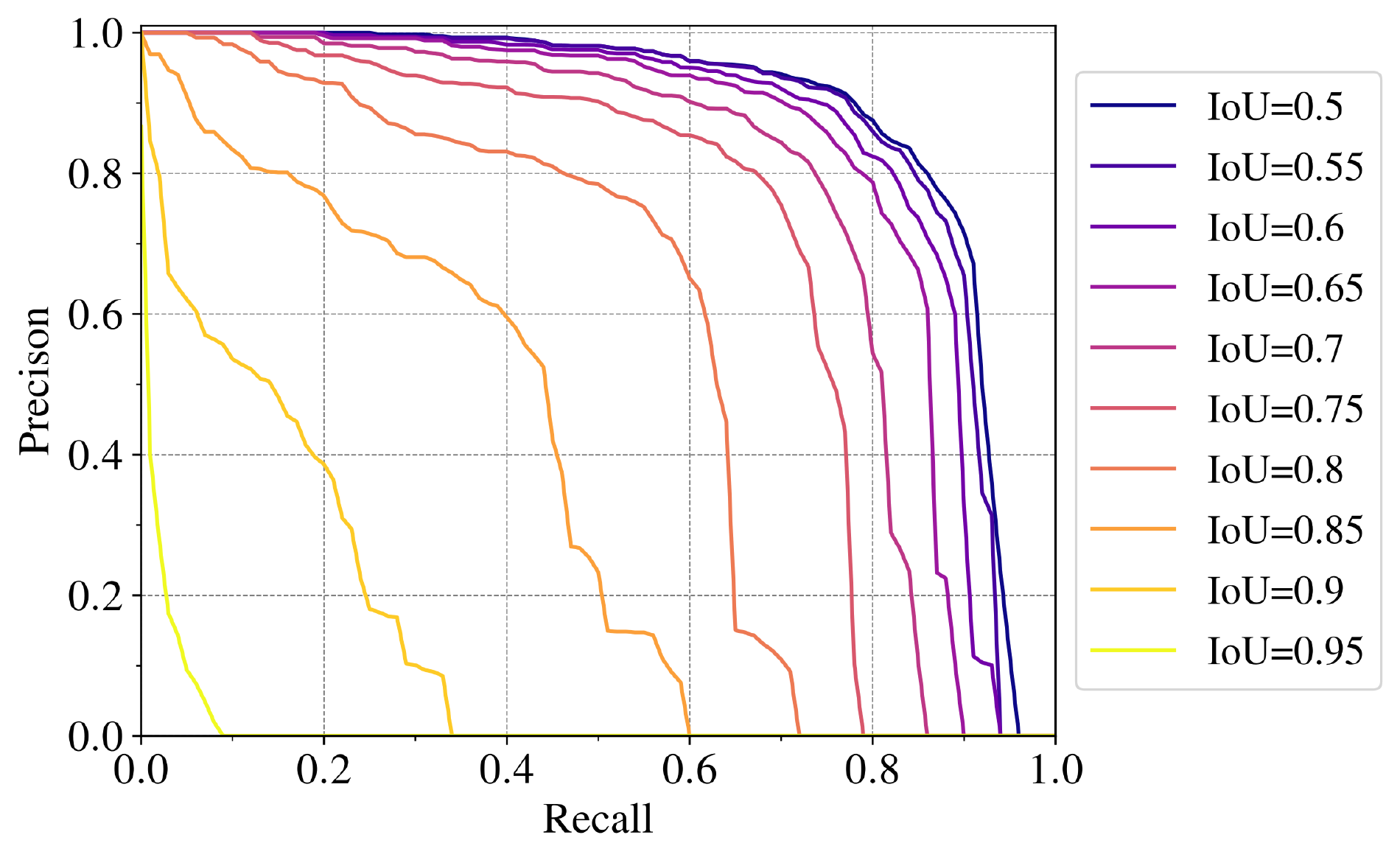}}
			\caption{Precision-recall curves of predicted bounding box on trees. For each \ac{IoU} threshold, we observe an expected phenomenon where the model's precision and recall diminish with each increase in \ac{IoU} threshold. In our case, the precision and recall values drop steeply for \ac{IoU}s above 0.75. 
				These results are from our best performing model, Cascade Mask R-CNN DB-Swin-S.}
			\label{fig:pr_curves}
		\end{figure}

	\textbf{Bounding box precision recall:}
		The quality of our tree detector can be measured by its ability to find all ground-truth trees (\ac{FN}=0, high recall), while identifying only relevant trees (\ac{FP}=0, high precision)~\citep{padilla2021comparative}. 
		% Concerning its applicability to autonomous tree felling operations, 
		% In case a \ac{FP} or \ac{FN} happens during forestry operations, there is a risk of decreasing productivity or damaging equipment.
		To quantitatively analyze this criterion, we show the precision $\times$ recall curve for bounding box detection in Figure~\ref{fig:pr_curves}.
		A large area under the curve tends to indicate both high precision and high recall.
		We observe that predictions with an \ac{IoU}\,=\,0.5 and recalls of up to 60\,\% obtain close to 100\,\% of precision.
		By incrementally increasing \ac{IoU}, we see a significant decrease when \ac{IoU} $> 0.75$, and this indicates that most bounding box predictions are within \ac{IoU}\,=\,[0.5, 0.75].
		Hence, bounding box predictions from our model can achieve high precision and recall rates at the expense of a lesser \ac{IoU}. 
		This is interesting, because our keypoint estimation method does not directly depend on bounding box localization accuracy.
		In other words, our method benefits from a certain margin of error on bounding box predictions without it affecting keypoint accuracy.

\begin{figure}[h!]
			\centering
			\centering{\includegraphics[width=0.45\linewidth]{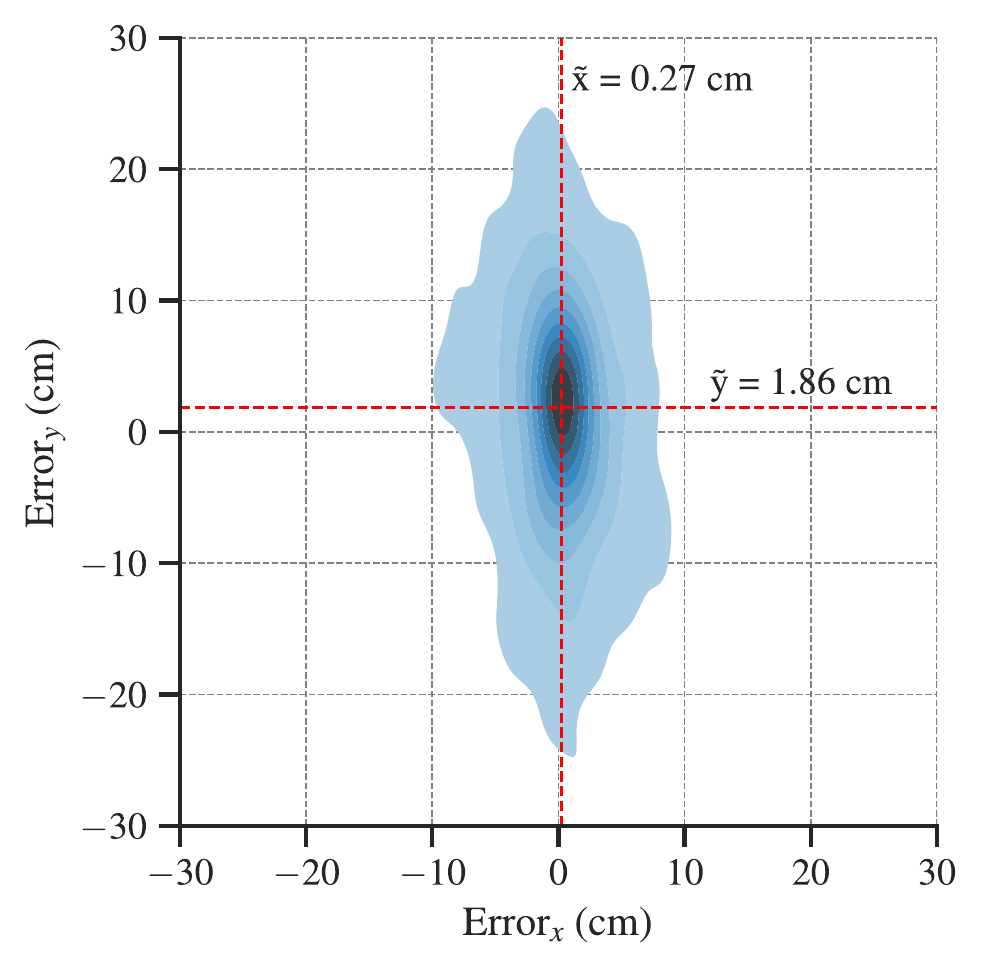}}
			\caption{Distribution of felling cut error. The lateral error median ($\tilde{x}$) is mostly centered at the origin. Whereas, the vertical error median ($\tilde{y}$) has a positive offset, and this is possibly caused by the presence of foliage in the lower part of the tree base. }
			\label{fig:fc_dist}
		\end{figure}

	\textbf{Felling cut position:} 
		The qualitative keypoint estimation results can be observed in Figure~\ref{fig:detection_examples}.
		To obtain the error in physical units, we simply used the associated depth image gathered with the ZED stereo camera to compute the 3D spatial position of the estimated keypoint versus its ground truth. 
		From Table~\ref{tab:mainResults}, we can observe that using keypoints, our models can estimate the felling cut position with an average error less than 7\,cm. 
		This accuracy is, at the moment, borderline for autonomous tree felling operations, since \citet{lindroos2015estimating} estimate that accuracy in the range of cm to mm is essential for efficient autonomous crane movement.
		Fortunately, we show in Figure~\ref{fig:fc_dist} that this error, with the image frame as referential, is mostly distributed vertically ($y$) rather than horizontally ($x$), with the $y$ error distribution twice as wide as the $x$ error.
		In practice, this means that the felling head would grip the tree in its center, but at a lower or higher position than its target height. 
		Intuitively, the vertical error can be attributed to the difficulty of estimating a height from the ground.
		Moreover, scenes with dense understorey foliage blocking tree base visibility lead to keypoint predictions that are higher up the trunk.
		This phenomenon can be observed in Figure~\ref{fig:fc_dist}, where the vertical error is more densely located in the $y$ half-plane with a median offset of +1.86\,cm.
		Notably, this vertical error $y$ can be partially mitigated by tapping the head of a feller buncher on the ground, and then pressing forward~\citep{ireland2009CCF}. 
		% *this is because image coordinates origin is top left, so high keypoints - lower gt keypoint will give negative error

\begin{figure}[h!]
			\centering
			\centering{\includegraphics[width=0.5\linewidth]{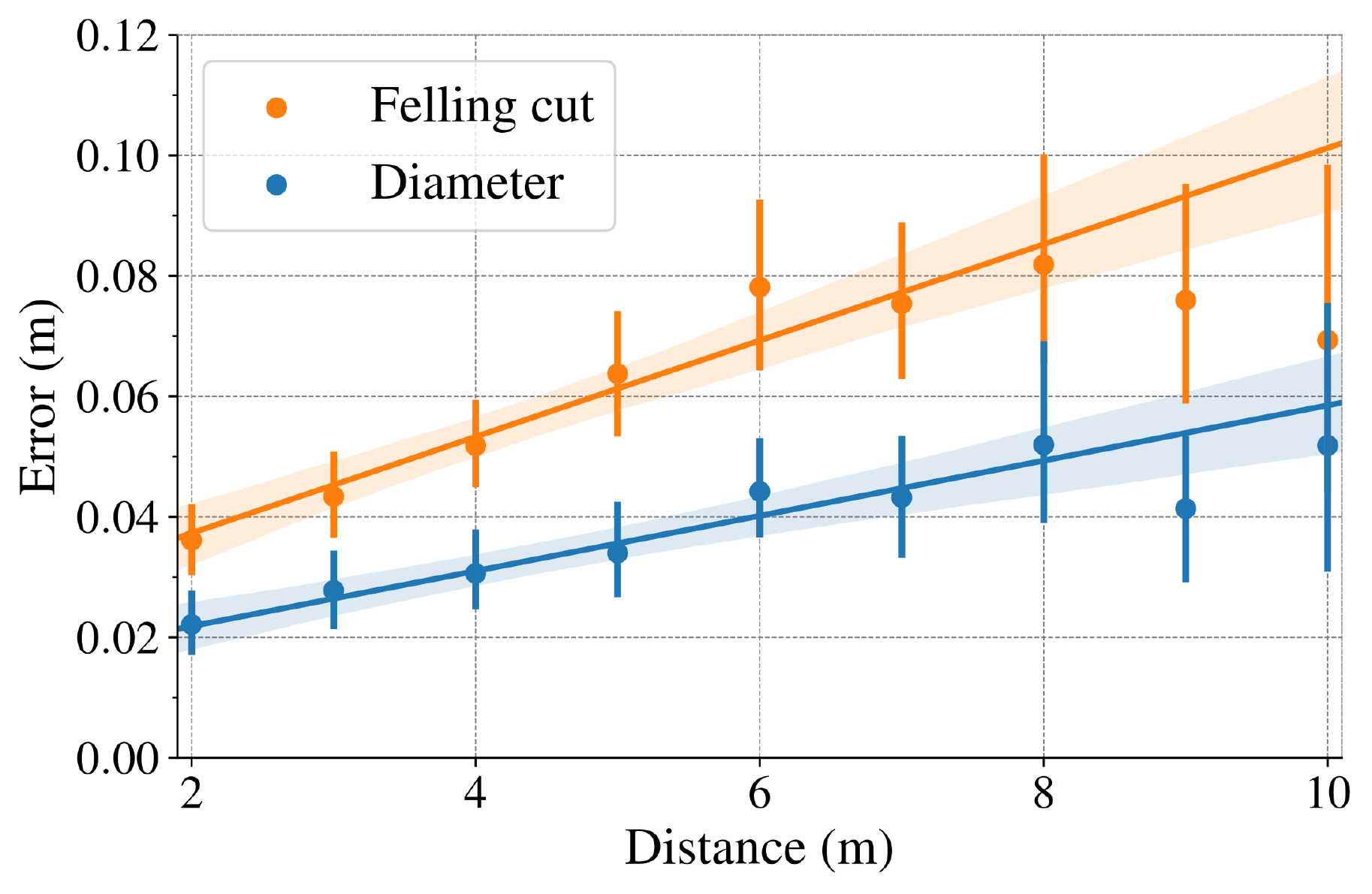}}
			\caption{Diameter and felling cut errors relative to tree distance from the camera. 
				We limit ourselves to a range of 10\,m, corresponding to the typical reach of machines~\citep{lindroos2015estimating}.
				Instances further than that are ignored. Translucent bands around the regression lines correspond to a 95\,\% confidence interval.}
			\label{fig:diameter_fc_func_depth}
		\end{figure}

	\textbf{Diameter estimation:}
		From Table~\ref{tab:mainResults}, the diameter estimation error is well under 4\,cm.
		The diameter error is almost half the size of the felling cut error (7\,cm). We believe the diameter estimation is facilitated by its relative independence from the height at which it is measured.
		We can also see in Figure~\ref{fig:diameter_fc_func_depth} that the diameter estimation error increases in relation to the distance between the camera and tree.
		This comes from the fact that keypoint estimation is done in pixel space, so the error grows linearly as a function of distance when converted to meters.
		Overall, the accuracy of our diameter estimation method is comparable to similar research using lidar, such as \citet{tremblay2020automatic} who report 2.04\,cm \ac{DBH} error for well-spaced trees and 3.45\,cm in dense forests.
		Nevertheless, a better accuracy would lower the risks of impairing control algorithms and damaging the machine or trees~\citep{lindroos2015estimating}. 
		Improving keypoint accuracy is important for autonomous tree felling operations, where accurate keypoint estimation translates to moving the harvester's head at the correct felling cut position, with the gripper opened wide enough and aligned with the tree's inclination.

	\subsection{Domain transfer}
	\label{subsection:Transfer}
		So far, the experiments suggest that deep learning approaches perform well for tree detection.
		Using both \PortugalDB and \CanaTreeDB datasets, we evaluate the domain transfer performance by training on the source domain and directly testing on the target domain.

		\begin{table*}[h!]
			\centering
			\caption{Domain transfer comparison between \CanaTreeDB and \PortugalDB{} datasets. Training is done on the source domain and directly tested on the target domain, without any fine-tuning. Results are obtained with our best architecture, Cascade Mask R-CNN DB-Swin-S, pre-trained on \SynthTreeDB. }\label{tab:crossDomain}
			\makebox[\linewidth]{
			% Scale down table to \textwidth
			\resizebox{16.4cm}{!}{%
			\begin{threeparttable}
				\begin{tabular}{c|c|cccc|cccc}
					\toprule%
					\multirow{2}{2.3cm}{Transfer Source$\rightarrow$Target} & \multirow{2}{1.8cm}{\# Training images} & \multicolumn{4}{@{}c@{}}{Source Domain} & \multicolumn{4}{@{}c@{}}{Target Domain} \\\cmidrule{3-10}
					& & AP$^{bb}_{50}$ & AP$^{bb}_{50:95}$ & AP$^{seg}_{50}$ & AP$^{seg}_{50:95}$ & AR$^{bb}_{50}$ & AR$^{bb}_{50:95}$ & AR$^{seg}_{50}$ & AR$^{seg}_{50:95}$ \\ 
					\midrule
					% YOLOv4 & \multirow{5}{*}{Port $\rightarrow$ Can} & 14000 &  &  &  &  &  &   \\ \cmidrule{1-1}
					\multirow{4}{*}{Port$\rightarrow$Can} & 100 & 76.7$^{\downarrow18.8}$ & 39.2$^{\downarrow36.7}$ & 85.6$^{\downarrow12.2}$ & 51.3$^{\downarrow30.6}$ & 72.1$^{\uparrow0.1}$ & 35.6$^{\downarrow1.8}$ & \textbf{82.0$^{\uparrow6.8}$} & 45.6$^{\uparrow3.0}$ \\ 
					& 1000 & 88.9$^{\downarrow6.6}$ & 59.6$^{\downarrow16.3}$ & 94.9$^{\downarrow2.9}$ & 68.9$^{\downarrow13.0}$ & \textbf{73.6$^{\uparrow1.6}$} & \textbf{42.3$^{\uparrow4.9}$} & 79.9$^{\uparrow4.7}$ & \textbf{49.9$^{\uparrow7.3}$} \\
					& 5000 & 93.3$^{\downarrow2.2}$ & 69.5$^{\downarrow6.4}$ & 96.8$^{\downarrow1.0}$ & 69.7$^{\downarrow12.2}$ & 68.9$^{\downarrow1.1}$ & 31.8$^{\downarrow5.6}$ & 74.8$^{\downarrow0.4}$ & 37.2$^{\downarrow5.4}$ \\
					& 14000 & \textbf{95.5} & \textbf{75.9} & \textbf{97.8} & \textbf{81.9} & 72.0 & 37.4 & 75.2 & 42.6 \\ \cmidrule{1-10}
					Can$\rightarrow$Port & 60 & 90.4 & 64.1 & 94.5 & 70.4 & 49.1 & 20.0 & 69.2 & 34.0 \\ 
					\bottomrule
				\end{tabular}
				% \begin{tablenotes}[para]
				%   \item[]Note: Entries are coefficients from a probit regression model. Robust standard errors in parentheses.
				%   \item[***] $p < 0.01$,
				%   \item[**] $p < 0.05$,
				%   \item[*] $p < 0.1$, two-tailed test.
				% \end{tablenotes}
			\end{threeparttable}
			}
			}
		\end{table*}

\begin{figure}[h!]
			\centering
			\includegraphics[width=1.0\linewidth]{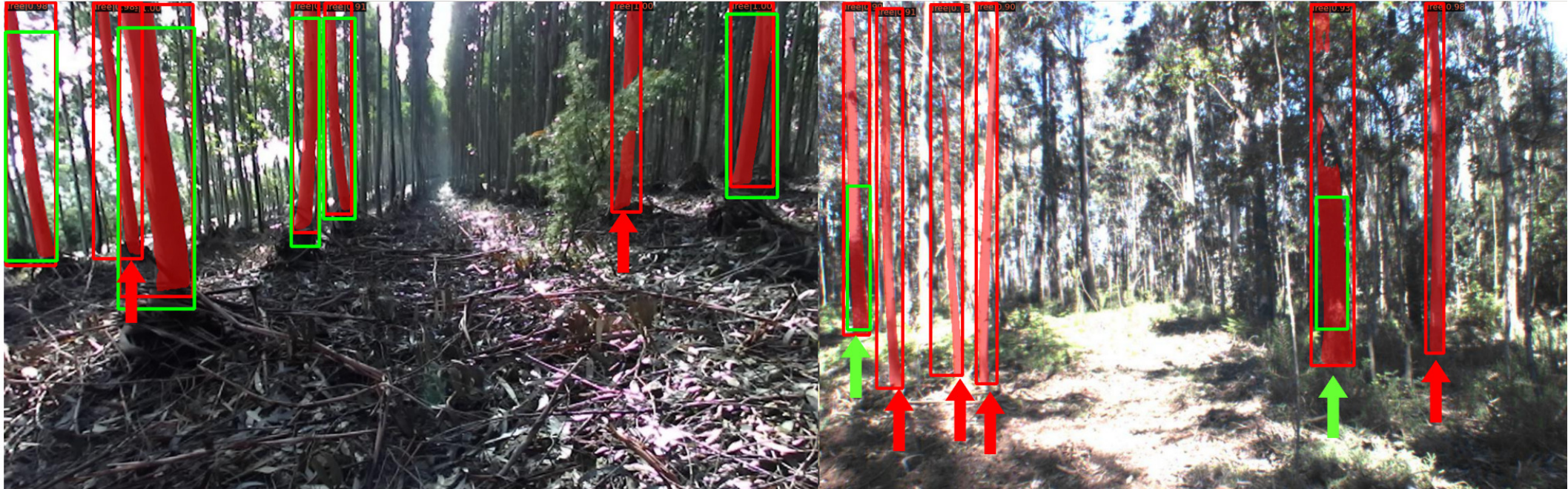}
			\caption{Domain transfer from \CanaTreeDB to \PortugalDB dataset and examples of annotation disparities. Ground truth annotations are in green and predictions by Cascade Mask R-CNN DB-Swin-S trained on \CanaTreeDB are in red, and include the predicted segmentation masks. \textbf{Left} red arrows point to trees that should be annotated, and \textbf{right} green arrows show bounding box annotations so small that they are counted as false negatives (\ac{IoU} $<$ 0.5). This leads to a decrease in the reported \ac{AP} and \ac{AR} results. 
			}
			\label{fig:PortugalInconsistancy}
		\end{figure}

	\textbf{Tree detection generalization:}
		From the results in Table~\ref{tab:crossDomain}, domain transfer is underwhelming, averaging a 22\,\% \ac{AP}$^{bb}_{50}$ and 17\,\% \ac{AR}$^{bb}_{50}$ decrease between the source and target domain.
		For the Can$\rightarrow$Port transfer, the \ac{AP}$^{bb}_{50}$ gap exceeds 40\,\%, which is odd since it qualitatively performs well (see Figure~\ref{fig:PortugalInconsistancy}).   
		Since a straightforward comparison of network performance between datasets produced by different people is hazardous, the reported results are probably lower than they should be.
		For instance, the process of image collection and annotation can vary.
		In our case, we observed that the bounding box annotations are based on different criteria such as minimum tree size, tree distance, and bounding box height.
		In \CanaTreeDB, the bounding boxes include every visible stem pixel up to the tree top, while the bounding boxes in \PortugalDB dataset are shorter (see Figure~\ref{fig:PortugalInconsistancy}).
		Since bounding box size directly impacts \ac{IoU}s, these annotation disparities can significantly decrease \ac{AP} and \ac{AR} results. 
		Moreover, there seems to be missing tree annotations in some of \PortugalDB images, which further cause a drop in \ac{AP} for the Can$\rightarrow$Port transfer.

\begin{figure}[h!]
			\centering
			\includegraphics[width=0.7\linewidth]{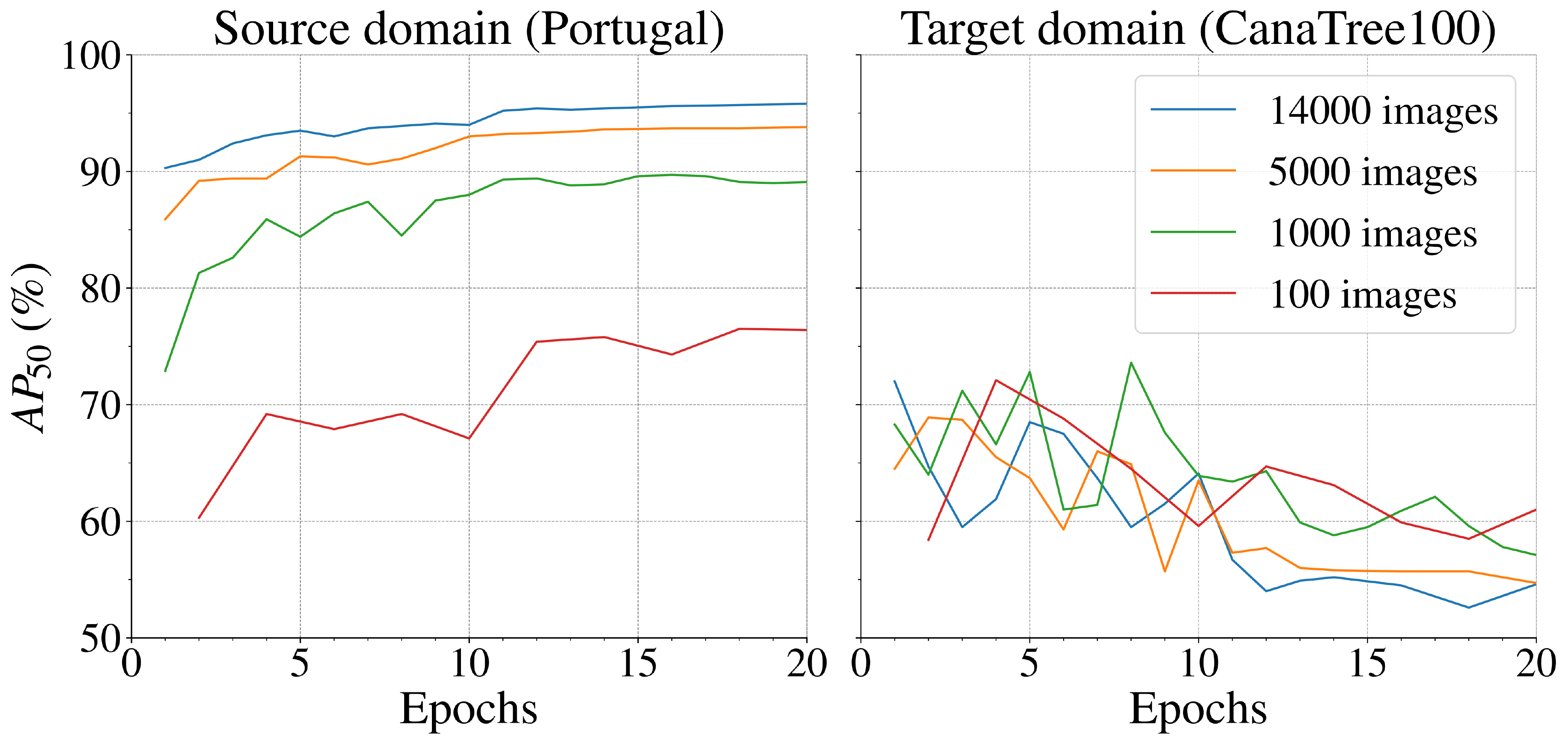}
			\caption{Impact of the number of training epochs and source dataset sizes on generalization, when going from \PortugalDB to \CanaTreeDB.
				On the \PortugalDB source domain, test performance increases with the number of training images and epochs, as seen on the left.
				However, when evaluated on \CanaTreeDB (on the right), we observe the opposite: models that were trained for few epochs generalize better. 
				Moreover, we can see that using more images in the source domain does not improve generalization results.
				Beyond 10 training epochs, performances degrade when passing from 100 to 14000 images.
			}
			\label{fig:cross_domain_results}
		\end{figure}

	\textbf{Overfitting:}
		When training on \PortugalDB as a source domain, we notice a significant overfitting effect across domains.
		As can be seen in Figure~\ref{fig:cross_domain_results}, performance on the \CanaTreeDB dataset degrades beyond a handful of epochs, while it keeps improving on the \PortugalDB source domain.
		Moreover, we observe that the number of training images from \PortugalDB has a significant impact on source domain performances, but this is not observed on the target domain.
		Indeed, training on 100 or 14000 images from \PortugalDB yields no significant improvement when testing on \CanaTreeDB as a target domain.
		This might be due to the strong visual correlation between samples in the \PortugalDB dataset, as they come from a limited set of videos (we approximate less than 6).
		In comparison, images from our \CanaTreeDB are extracted from 33 videos.
		Note that when using \CanaTreeDB as a source to test on \PortugalDB, the generalization gap was more or less constant across the epochs from start to end, and it did not change by more than 3~\%, in comparison to over 20~\% for \PortugalDB.

	\textbf{Generalization difficulties:}
		The underwhelming domain transfer performance can partly be attributed to annotation disparities between datasets. Yet, the current experiment still indicates a problematic decrease in perception performance when changing domain. 
		These domain changes are caused by different camera sensor type, the camera lens, variation in forest's vegetation due to geography, etc.
		Drastic domain changes can lead to generalization difficulties and slow down the adoption of automation in forest operations.
		This is particularly relevant for international equipment manufacturers, as their machinery is expected to operate in a variety of forest sites and continents.
		This short experiment shows that there might be a need to sample train images from forests that are in the geographical areas where these machines are to be deployed until better domain adaptation techniques are developed.

	\subsection{Further Experiments}
	\label{subsection:Ablation}
	
	\textbf{Pre-training on synthetic images:}
		Synthetic images have the potential to improve detection results without incurring major costs like time or human resources.
		In this experiment, we quantitatively measure the usefulness of our synthetic dataset (\SynthTreeDB) for tree detection and whether it benefits transfer learning.
		Four pre-training strategies are tested using our Cascade Mask R-CNN DB-Swin-S pre-trained on COCO.
		In the first case, no subsequent pre-training is performed aside from the initial COCO one.
		For the other three, we use either \PortugalDB, \SynthTreeDB or both.
		Then, the models are fine-tuned on the downstream tasks of \CanaTreeDB.
		
		\begin{table*}[h!]
			\centering
			\caption{Impact of the datasets used for pre-training. Cascade Mask R-CNN DB-Swin-S model is pre-trained on one or more datasets (C: COCO, S:\SynthTreeDB, P:\PortugalDB), then fine-tuned and evaluated on \CanaTreeDB. Pre-training on the synthetic dataset (S) improves detection performances, superseding \PortugalDB.} \label{tab:abla_synth_depth}
			\begin{threeparttable}
				\begin{tabular}{c|cccc|cccc}
					\toprule%
					Pre-training & AP$^{bb}_{50}$ & AP$^{bb}_{50:95}$ & AP$^{seg}_{50}$ & AP$^{seg}_{50:95}$ & AR$^{bb}_{50}$ & AR$^{bb}_{50:95}$ & AR$^{seg}_{50}$ & AR$^{seg}_{50:95}$ \\ 
					\midrule
					C & 83.5$_{\pm 1.2}$ & 59.3$_{\pm 2.3}$ & 81.0$_{\pm 2.1}$ & 54.2$_{\pm 1.6}$ & 87.6$_{\pm 2.1}$ & 65.7$_{\pm 2.5}$ & 84.9$_{\pm 2.9}$ & 60.1$_{\pm 2.0}$ \\ 
					C+P & 86.6$^{\uparrow3.1}_{\pm 1.4}$ & 63.2$^{\uparrow3.9}_{\pm 1.3}$ & 84.7$^{\uparrow3.7}_{\pm 1.6}$ & 57.5$^{\uparrow3.3}_{\pm 0.8}$ & 90.9$^{\uparrow3.3}_{\pm 2.1}$ & 69.7$^{\uparrow4.0}_{\pm 1.0}$ & 88.7$^{\uparrow3.8}_{\pm 1.1}$ & 63.0$^{\uparrow2.9}_{\pm 0.7}$  \\ 
					C+S & \textbf{90.4$^{\uparrow6.9}_{\pm 2.4}$} & 64.1$^{\uparrow4.8}_{\pm 1.4}$ & \textbf{87.2$^{\uparrow6.1}_{\pm 1.7}$} & \textbf{60.0$^{\uparrow5.8}_{\pm 1.4}$} & \textbf{94.5$^{\uparrow6.9}_{\pm 1.9}$} & 70.4$^{\uparrow4.7}_{\pm 1.1}$ & \textbf{91.5$^{\uparrow6.6}_{\pm 1.4}$} & \textbf{65.2$^{\uparrow5.1}_{\pm 1.1}$} \\
					C+S+P & 89.5$^{\uparrow6.0}_{\pm 1.0}$ & \textbf{65.2$^{\uparrow5.9}_{\pm 0.8}$} & 86.3$^{\uparrow5.3}_{\pm 1.7}$ & 59.9$^{\uparrow5.7}_{\pm 1.2}$ & 93.8$^{\uparrow6.2}_{\pm 1.2}$ & \textbf{71.4$^{\uparrow4.7}_{\pm 0.5}$} & 90.8$^{\uparrow5.9}_{\pm 1.8}$ & \textbf{65.2$^{\uparrow5.1}_{\pm 1.1}$} \\
					% \midrule
					% Depth & Synth & 79.9 & 45.5 & 70.1 & 35.4 & 87.8 & 53.7 & 76.8 & 41.6 \\ 
					\bottomrule
				\end{tabular}
				% \begin{tablenotes}[para]
				%   \item[]Note: Entries are coefficients from a probit regression model. Robust standard errors in parentheses.
				%   \item[***] $p < 0.01$,
				%   \item[**] $p < 0.05$,
				%   \item[*] $p < 0.1$, two-tailed test.
				% \end{tablenotes}
			\end{threeparttable}
		\end{table*}
		
		From Table~\ref{tab:abla_synth_depth}, we can see that the most effective pre-training strategy is based on synthetic images, outperforming COCO-only and \PortugalDB by at least 3.8\,\% AP$^{bb}_{50}$, and 3.6 AR$^{bb}_{50}$.
		It also shows that adding images from the \PortugalDB dataset does not provide significant improvements when synthetic images are already used.
		This clearly indicates that pre-training on a large-scale, densely annotated synthetic dataset, such as \SynthTreeDB, is a great starting point to learn powerful and robust representation for tree detection.

	\textbf{Robustness to occlusion:}
		% \label{subsubsection:occlusion}
		In forests, computer vision algorithms are subject to many distractors such as branches, bushes or the machinery itself causing occlusion. 
		Therefore, we conduct a pixel-level analysis to measure the robustness of our method when faced to different levels of occlusion.
		
		\begin{figure}[hbt!]
			\centering
			\includegraphics[width=0.6\linewidth]{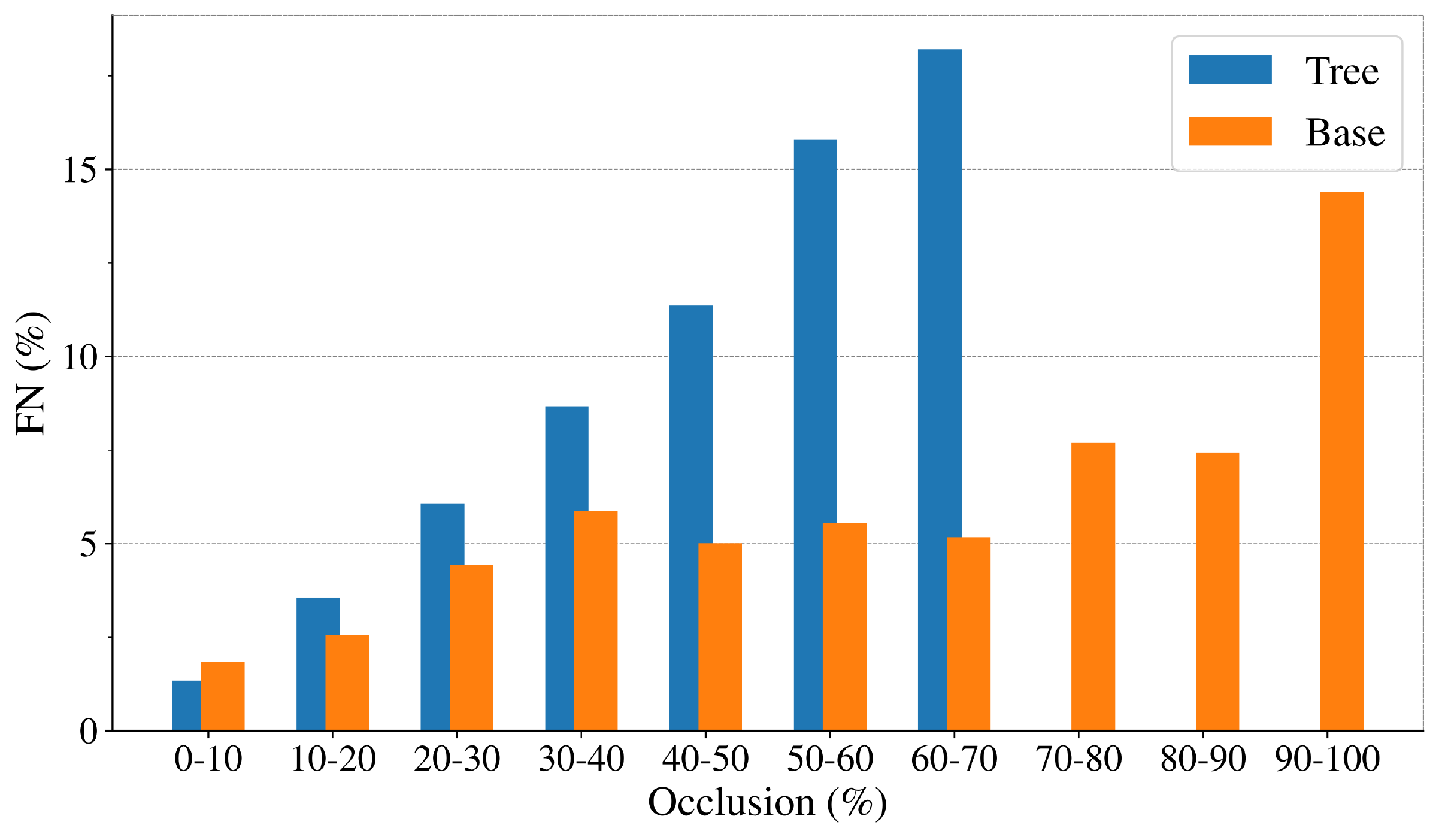}
			\caption{Impact of occlusion on tree detection, for synthetic images, in terms of \ac{FN} rates. Occlusion of the tree base (0 to 30\,cm above ground) has significantly less impact than for the entire trees.
				Without surprise, the lowest \ac{FN} rate corresponds to the lowest occlusion level.
			}
			\label{fig:occlusion_fn_hist}
		\end{figure}
		
		From Figure~\ref{fig:occlusion_fn_hist}, we observe an increasing rate of \ac{FN} (missed detections) relative to the percentage of occlusion. 
		For tree occlusion, the \ac{FN} rate grows linearly, but for the tree base, it spikes when the base is over 90\,\% occluded.
		This indicates that tree base occlusion is less impactful than tree occlusion itself, which can be useful in the presence of dense understorey foliage. 
		In fact, although at least 30\,\% of each tree is visible to the camera (our annotation threshold), there is only 14\,\% \ac{FN} in cases where the tree base area is over 90\,\% occluded.
		In light of these findings, it is without surprise that the ideal conditions for tree detection are well-maintained forests with minimal occlusion.
		% because the lowest \ac{FN} rate corresponds to the lowest occlusion level
		Even though this experiment is conducted on synthetic images, due to the difficulty of measuring occlusion in real images, it provides key insights on how robust our method can be when faced with occlusion in real images.

\begin{figure}[hbt!]
			\centering
			\includegraphics[width=0.6\linewidth]{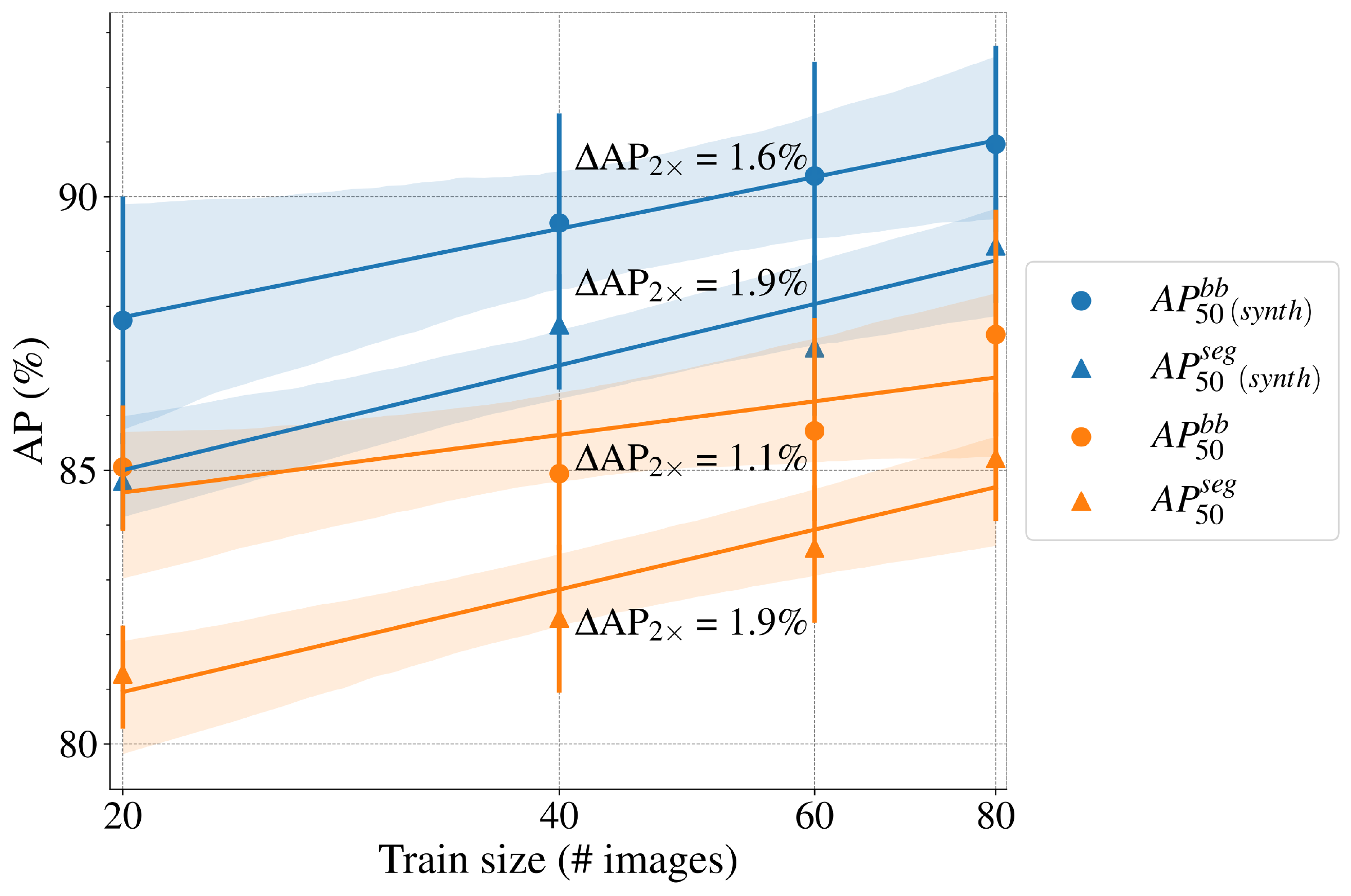}
			\caption{Performances scale with the train set size. Reported results are with and without pre-training on synthetic data, fine-tuned and tested on \CanaTreeDB. The slope of $\Delta$\ac{AP} represents the change in \ac{AP} each time the train set size doubles. Translucent bands around the regression lines correspond to a 95\,\% confidence interval.
			}
			\label{fig:train_size_regression}
		\end{figure}

	\textbf{Dataset size:}
		Deep learning is well known for scaling with large amounts of data.
		This experiment seeks to quantitatively estimate the performance improvements if we were to increase the number of annotated samples in \CanaTreeDB. 
		To do so, we vary the train set size between 20 and 80 images, keeping the same 20 image test set for each fold.
		As shown by the logarithmic abscissa scale in Figure~\ref{fig:train_size_regression}, we found that detection improvement follows a power law.
		Each time the number of training images doubles, we gain between 1.1 to 1.6\,\% AP$^{bb}_{50}$, and 1.9\,\% AP$^{seg}_{50}$. 
		Moreover, by pre-training \SynthTreeDB , we remark a near-constant gain of 4.0\,\% AP$^{bb}_{50}$ and 4.0\,\% AP$^{seg}_{50}$ compared to the model pre-trained solely on COCO.
		This further consolidates the fact that our synthetic forest dataset can be leveraged to achieve better performances, compared to training a limited number of real images only.
		Markedly, Figure~\ref{fig:train_size_regression} illustrates that pre-training on synthetic images and fine-tuning on only 20 images outperforms the model pre-trained on COCO-only and fine-tuned on 80 (four times more) real images. 
		In a low-data regime like ours, this can allow us to partially fill the data gap and avoid hand annotation. 
		
	\textbf{Keypoint branch improvements}
		In order to implement autonomous movements, accurate end-effector target position is desirable.
		Here, we seek to reduce the felling cut error obtained with the CNN keypoint head by replacing it with TransPose~\citep{yang2021transpose}, a transformer-based keypoint head.
		The performance comparison between these two keypoint head architectures is presented in Table~\ref{tab:transpose_results}, where we observe a 20\,\% (1 cm) error reduction for felling cut predictions made with the TransPose approach. 
		
		\begin{table}[hbt!]
			\centering
			\caption{Comparison between network head architectures for keypoint predictions. %Depth is the feed-forward resolution $\times$ number of convolution layer (CNN) or encoder layer (TransPose).
				TransPose achieves more accurate keypoint predictions.
				However, both architectures fail to scale with depth. Model is Mask R-CNN Swin-S pre-trained on the synthetic dataset and fine-tuned on \CanaTreeDB dataset.}\label{tab:transpose_results}
			\begin{threeparttable}
				\begin{tabular}{c|c|ccc}
					\toprule%
					\multirow{2}{1.8cm}{Keypoint Head} & \multirow{2}{3cm}{Architecture (channel$\times$depth)}  & \multicolumn{3}{@{}c@{}}{Keypoint error} \\\cmidrule{3-5}
					& & dia$_{(cm)}$ & fc$_{(cm)}$ & inc$_{(^{\circ})}$) \\ 
					\midrule
					\multirow{3}{1.8cm}{CNN} & $256\times8$  & 3.1 & 6.3 & 1.7 \\
					& $512\times8$  & 3.3 & 6.5 & 1.9  \\ 
					& $256\times16$ & 2.9 & 6.5 & \textbf{1.4} \\
					\midrule
					\multirow{4}{1.8cm}{TransPose} & $512\times4$  & 3.0 & 5.6 & 1.87  \\
					& $512\times6$   & \textbf{2.8} & 5.6 & 1.6  \\
					& $1024\times6$ & 3.0 & 5.6 & 1.7   \\
					& $512\times8$ & 3.0 & \textbf{5.4} & 1.5   \\
					\bottomrule
				\end{tabular}
				% \begin{tablenotes}[para]
				%   \item[]Note: Entries are coefficients from a probit regression model. Robust standard errors in parentheses.
				%   \item[***] $p < 0.01$,
				%   \item[**] $p < 0.05$,
				%   \item[*] $p < 0.1$, two-tailed test.
				% \end{tablenotes}
			\end{threeparttable}
		\end{table}
		
\begin{figure}[hbt!]
			\centering
			\includegraphics[width=0.65\linewidth]{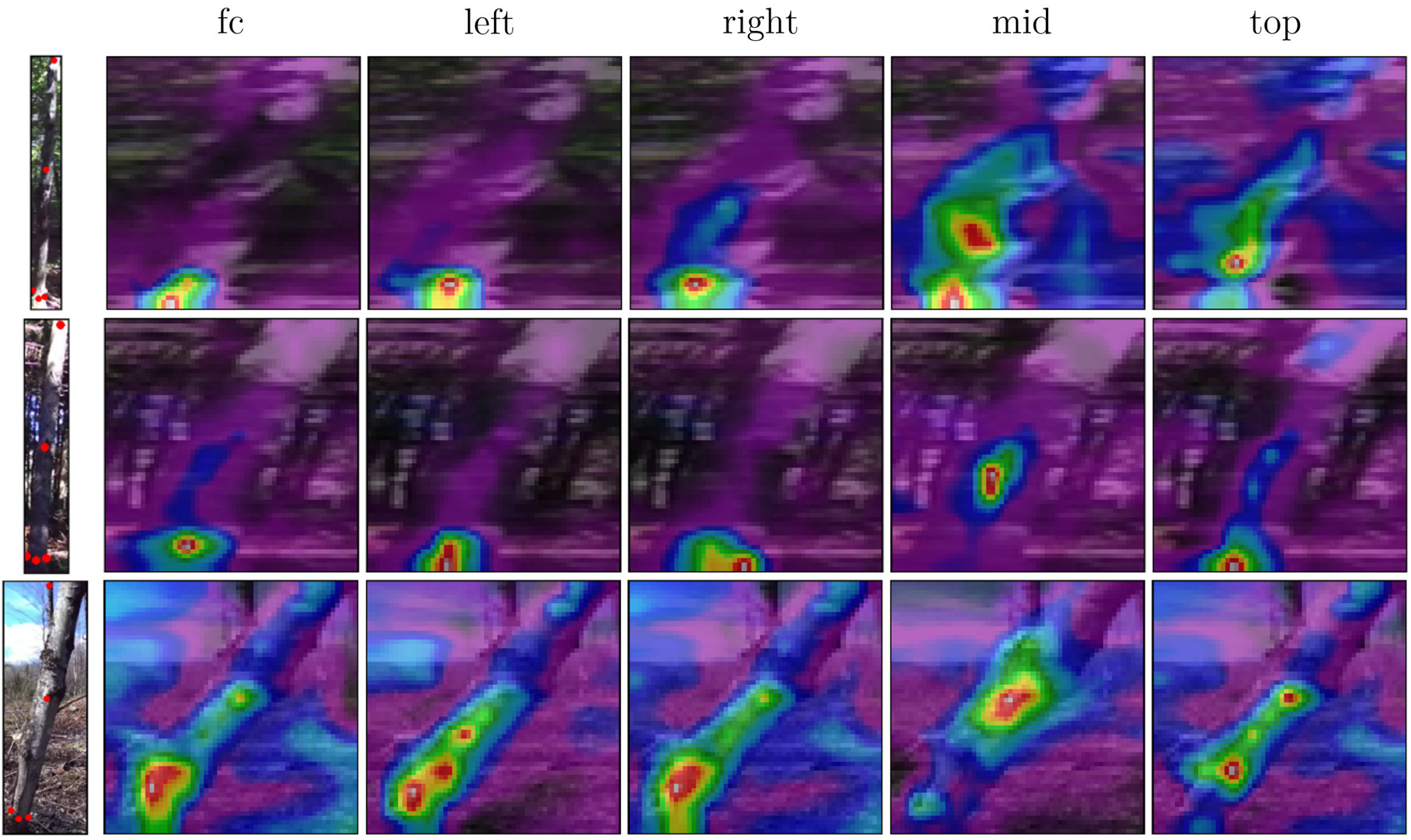}
			\caption{Heatmap of attention scores for a particular keypoint position. The felling cut, left and right diameter keypoints have obvious local dependencies, whereas mid and top keypoints exploit long range image cues. Red areas indicate higher attention scores.
				The rectangular images on the left are stretched into a square one by the RoIAlign layer.}
			\label{fig:attention_map}
		\end{figure}

		However, we did not see the performance scale with increased depth, but we observe a saturation or decrease in performance. 
		This indicates that the attention mechanism at the core of transformers is responsible for the increased accuracy and is indeed effective at capturing pertinent information from different image locations.
		For instance, given a query location (keypoint) and an attention map, one can explicitly reveal the dependency areas on which each keypoint prediction is based.
		As can be visually confirmed in Figure~\ref{fig:attention_map}, keypoints in the tree base area have obvious dependencies on visual cues from their local area, while keypoints higher up the stem rely on long-range cues. 
		This potentially allows the model to localize partially obstructed keypoints by looking for more significant visual cues, thus reducing its reliance on the occluded pixels.

\section{Conclusion and Future Works}\label{sec:Conclusion}
	In this paper, we demonstrate that supervised end-to-end deep learning can successfully detect trees in forested environments.
	For this, we tackle the problematic lack of datasets by creating and publicly releasing two novel forest image datasets --- one synthetic (\SynthTreeDB) and one real (\CanaTreeDB).
	Based on the intuitive paradigm of pre-training on a large synthetic image dataset followed by fine-tuning on a real image dataset, we achieve promising results in all three major perception tasks: 90.4\,\% tree detection precision, 87.2\,\% segmentation precision and cm accurate keypoint estimations.
	Notably, we show that diameter estimation using a vision-based approach achieves comparable results to methods based on lidar, while also providing felling cut location and tree inclination. 
	
	In terms of shortcomings, the limited capacity to generalize to other forest sites suggests that training on specific forest sites is required before practical applications.
	Given the relative ease of training, there is optimism in the scalability of our approach to dataset size and model architecture.
	We show that detection performances scale with dataset size, follow a power law that increases by 1.1 to 1.6\,\% AP$^{bb}_{50}$, and 1.9\,\% AP$^{seg}_{50}$ every time the training set size doubles.
	In addition, changing the model backbone, architecture, or pre-training dataset can give direct performance gains.
	Again, these findings outline the need for much larger, rigorously annotated forest datasets. 
	Hopefully, the paper has sufficiently highlighted the pertinence of releasing datasets so others can benchmark on it and allow the community to clearly establish a state-of-the-art method for tree detection.
	
	Future works should investigate the creation of a larger real image dataset, but also ways to enhance model generalization to different forest sites, and conduct evaluations on independent test sets~\citep{diez2021deep}.
	Since the reliance on annotated datasets is taking a toll on progresses made in fields like forestry, which receive far less attention from the deep learning community, semi-supervised or self-supervised learning methods are worth investigating. 
	%ur tree detection system is ready for field deployments and experiments onboard a harvester which could advance the community. 

\bibliographystyle{unsrtnat}
\bibliography{references}  %%% Uncomment this line and comment out the ``thebibliography'' section below to use the external .bib file (using bibtex) .





% \end{thebibliography}

\end{document}